%% file: main.tex
\documentclass[a4paper]{article}
\pdfoutput=1
\usepackage{graphicx}
\usepackage{url}
\usepackage{amsmath,amsfonts,amssymb}
\usepackage{math}
\usepackage{algorithm}
\usepackage{algorithmic}
\usepackage{booktabs}
\usepackage[caption=false,font=footnotesize]{subfig}
\usepackage{multirow}
\usepackage[font=footnotesize,labelfont=bf]{caption}
\usepackage{ifpdf}
\usepackage{xcolor}

\newcommand{\reffig}[1]{Figure~\ref{#1}}
\newcommand{\refsec}[1]{Section~\ref{#1}}
\newcommand{\reftab}[1]{Table~\ref{#1}}
\newcommand{\refalg}[1]{Algorithm~\ref{#1}}
\newcommand{\refeq}[1]{(\ref{#1})}
\renewcommand{\th}{^\textrm{th}}

\renewcommand{\p}{\textrm{p}}

\usepackage{anysize}
\marginsize{2.5cm}{2.5cm}{2.5cm}{2.5cm}

\title{Vision-Guided Robot Hearing}
\author{Xavier Alameda-Pineda and Radu Horaud \vspace{0.4cm}\\ {\normalsize Perception Team, INRIA Grenoble
Rh\^one-Alpes} \\ {\normalsize655 Av. de l'Europe, 38334, Montbonnot, France} \\
{\normalsize\texttt{firstname.lastname@inria.fr}}}
\date{Submitted at International Journal of Robotics Research, Special Issue on Robot Vision}

\begin{document}
\maketitle
\begin{abstract}

Natural human-robot interaction in complex and unpredictable environments is one of the main research lines in robotics.
In typical real-world scenarios, humans are at some distance from the robot and the acquired signals are
strongly impaired by noise, reverberations and other interfering sources. In this context, the detection and
localisation of speakers plays a key role since it is the pillar on which several tasks (e.g.: speech recognition and
speaker tracking) rely. We address the problem of how to detect and localize people that are both seen and
heard by a humanoid robot. We introduce a hybrid deterministic/probabilistic model. Indeed, the deterministic component
allows us to map the visual information into the auditory space. By means of the probabilistic component, the visual
features guide the grouping of the auditory features in order to form AV objects. The proposed model and the
associated algorithm are implemented in real-time (17 FPS) using a stereoscopic camera pair and two microphones embedded
into the head of the humanoid robot NAO. We performed experiments on (i) synthetic data, (ii)
a publicly available data set and (iii) data acquired using the robot. The results we obtained validate the approach
and encourage us to further investigate how vision can help robot hearing.
\end{abstract}

\section{Introduction}
\label{sec:intro}

\input{introduction-radu}

\section{Related Work}
\label{sec:sota}
While vision and hearing have been mainly addressed separately, several behavioral, electrophysiological and imaging
studies \cite{Calvert04}, \cite{Ghazanfar06}, \cite{Senkowski08} postulate that the fusion of different sensorial
modalities is an essential component of perception. Nevertheless, computational models of audio-visual fusion and their
implementation on robots remain largely unexplored.
%

The problem of integrating data gathered with physically different sensors, e.g., cameras and microphones, is extremely
challenging. Auditory and visual sensor-data correspond to different physical phenomena which must be interpreted in a
different way. Relevant visual information must be inferred from the way light is reﬂected by scene objects and valid
auditory information must be inferred from the perceived signals such that it contains the properties of one or several
emitters. The spatial and temporal distributions of auditory and visual data are also very different. Visual data are
spatially dense and continuous in time. Auditory data are spatially sparse and intermittent since in a natural
environment there are only a few acoustic sources. These two modalities are perturbed by different phenomena such as
occlusions and self-occlusions (visual data) or ambient noise and echoic environments (auditory data).

Despite all these challenges, numerous researchers investigated the fusion of auditory and visual cues in
a variety of domains such as event classification \cite{Natarajan12}, speech recognition \cite{Barker09}, sound source
separation \cite{Naqvi10}, speaker tracking \cite{Hospedales08}, \cite{Gatica07} and speaker diarization
\cite{Noulas12}. However, these approaches are not suitable for robots either because the algorithmic
complexity is too high, or because methods use a distributed sensor network or because the amount of
training data needed is too high, drastically reducing the robots' adatableness. Unfortunately, much less effort has been
devoted to design audio-visual fusion methods for humanoid robots. Nevertheless, there are some interesting works
introducing methods specifically conceived for humanoid robots on speech recognition \cite{Nakadai04}, beat tracking
\cite{Itohara11}, \cite{Itohara12}, active audition \cite{Kim07} and sound recognition \cite{Nakamura11a}. All
these methods deal with the detection and localisation problem by using a combination of off-the-shelf algorithms,
suitable for humanoid robots. Albeit, all these approaches lack from a framework versatile enough to be used in other
situations than the ones they are specifically designed for.

Finding the potential speakers and assessing their speaking status is a pillar task, on which all applications mentioned
above rely. In other words, providing a robust framework to count how many speakers are in the scene, to localize them
and to ascertain their speaking state, will definitely increase the performance of many audio-visual perception
methods. This problem is particularly interesting in the case of humanoid robots, because the framework must be
designed for \textit{untethered} interaction using a set of \textit{robocentric sensors}. That is to say that the
cameras and microphones are mounted onto a robotic platform that freely interacts with the unconstrained AV events
(i.e., people). As a consequence, the use of any kind of distributed sensor network, e.g. close-range microphones and
speaker-dedicated cameras, is not appropriated. Likewise, the algorithms should be light enough to satisfy the
constraints associated to real-time processing with a humanoid robot.

The existing literature on speaker detection and localisation can be grouped into two main research lines. On one side,
many statistical non-parametric approaches have been developed. Indeed, \cite{Gurban06}, \cite{Besson08a} and
\cite{Besson08b} investigate the use of information theory-based methods to associate auditory and visual data in order
to detect the active speaker. Similarly, \cite{Barzelay07} proposes an algorithm matching auditory and visual onsets.
Even though these approaches show very good performance results, they use speaker/object dedicated cameras, thus
limiting the interaction. Moreover, the cited non-parametric approaches need a lot of training data. The outcome of
such training steps is also environment-dependent. Consequently, implementing such methods on mobile platforms results
in systems with almost no practical adaptability.

On the other side, several probabilistic approaches have been published. In \cite{Khalidov08}, \cite{Khalidov11a}, the
authors introduce the notion of conjugate GMM for audio-visual fusion. Two GMMs are estimated, one for each modality
(vision and auditory) while the two mixture parameter sets are constrained through a common set of \textit{tying
parameters}, namely the 3D locations of the AV events being sought. Recently in \cite{Noulas12}, a factorial HMM is
proposed to associate auditory, visual and audio-visual features. All these methods simultaneously detect and localize
the speakers but they are not suitable for real-time processing, because of their algorithmic complexity. \cite{Kim07}
proposed a Bayesian framework inferring the position of the active speaker and combining a sound source localisation
technique with a face tracking algorithm on a robot. The reported results are good in the case of one active speaker,
but show bad performance for multiple/far speakers. This is due to the fact that the proposed probabilistic framework is
not able to correctly handle outliers. In \cite{Alameda11}, the authors use a 1D GMM to fuse the auditory and visual
data, building AV clusters. The probabilistic framework is able to handle the outliers thanks to one of the
mixture components. However, the algorithm presented in the paper is not light enough for real-time processing.

Unlike these recent approaches, we propose a novel hybrid deterministic/probabilistic model for audio-visual detection
and localisation of speaking people. Up to the authors' knowledge, we introduce the very first model with the
following remarkable attributes all together: (i) theoretically sound and solid, (ii) designed to process robocentric
data, (iii) accommodating different visual and auditory features,  (iv) robust to noise and outliers, (v) requiring a
once-and-forever tiny calibration step guaranteeing the adaptability of the system, (vi) working on unrestricted indoor
environments, (vii) handling a variable number of people and (viii) implemented on a humanoid platform.

\section{A Hybrid Deterministic/Probabilistic Model}
\label{sec:problem}
We introduce a multimodal deterministic/probabilistic fusion model for audio-visual detection and localisation of
speaking people that is suitable for real-time applications. The algorithms derived from that hybrid model aim to count
how many speakers are there, find them in the scene and ascertain when they speak. In other words, we seek for the
number of potential speakers, $N\in\mathbb{N}$, their positions $\Svect_n\in\mathbb{S}$ ($\mathbb{S}\subset\mathbb{R}^3$
is the scene space) and their speaking state $e_n\in\{0,1\}$ (0 -- \textit{not speaking} and 1 -- \textit{speaking}).

In order to accomplish the detection and localization of speakers, auditory and visual features are extracted from the
raw signals (sound track and image flow), during a time interval $\Delta t$. We assume $\Delta t$ to be short enough
such that the speakers remain approximately in the same 3D location and long enough to capture small
displacements and oscillatory movements of the head, hands, torso and legs. The auditory and visual features extracted
during $\Delta t$ are denoted by $\amat=\left\{\avect_1,\ldots,\avect_k,\ldots,\avect_K\right\}\subset\mathbb{A}$ and by
$\vmat=\left\{\vvect_1,\ldots,\vvect_m,\ldots,\vvect_M\right\}\subset\mathbb{V}$ respectively, where $\mathbb{A}$
($\mathbb{V}$) is the auditory (visual) feature space.

We aim to solve the task from the auditory and visual observations. That is, we want to compute the
values of $N$, $\{\Svect_n\}_{n=1}^N$ and $\{e_n\}_{n=1}^N$, that best explain the extracted features $\amat$ and
$\vmat$. Therefore, we need a framework that encompasses all (hidden and observed) variables and that accounts for the
following challenges: (i)~the visual and auditory observations lie in physically different spaces with different
dimensionality, (ii) the object-to-observation assignments are not known in advance, (iii) both visual and auditory
observations are contaminated with noise and outliers, (iv) the relative importance of the two types of data is
unassessed, (v) the position and speaking state of the speakers has to be gauged and (vi) since we want to be able to
deal with a variable number of AV objects over a long period of time, the number of AV object that are effectively
present in the scene must be estimated.

We propose a hybrid deterministic/probabilistic framework performing audio-visual fusion, seeking for the desired
variables and accounting for the outlined challenges. On one hand, the deterministic components allow us to model those
characteristics of the scene that are known with precision in advance. They may be the outcome of a very accurate
calibration step, or the direct consequence of some geometrical or physical properties of the sensors. On the other
hand, the probabilistic components model random effects. For example, the feature noise and outliers, which is a
consequence of the contents of the scene as well as the feature extraction procedure.

\subsection{The Deterministic Model}
\begin{figure}
 \centering
\includegraphics[width=0.7\linewidth]{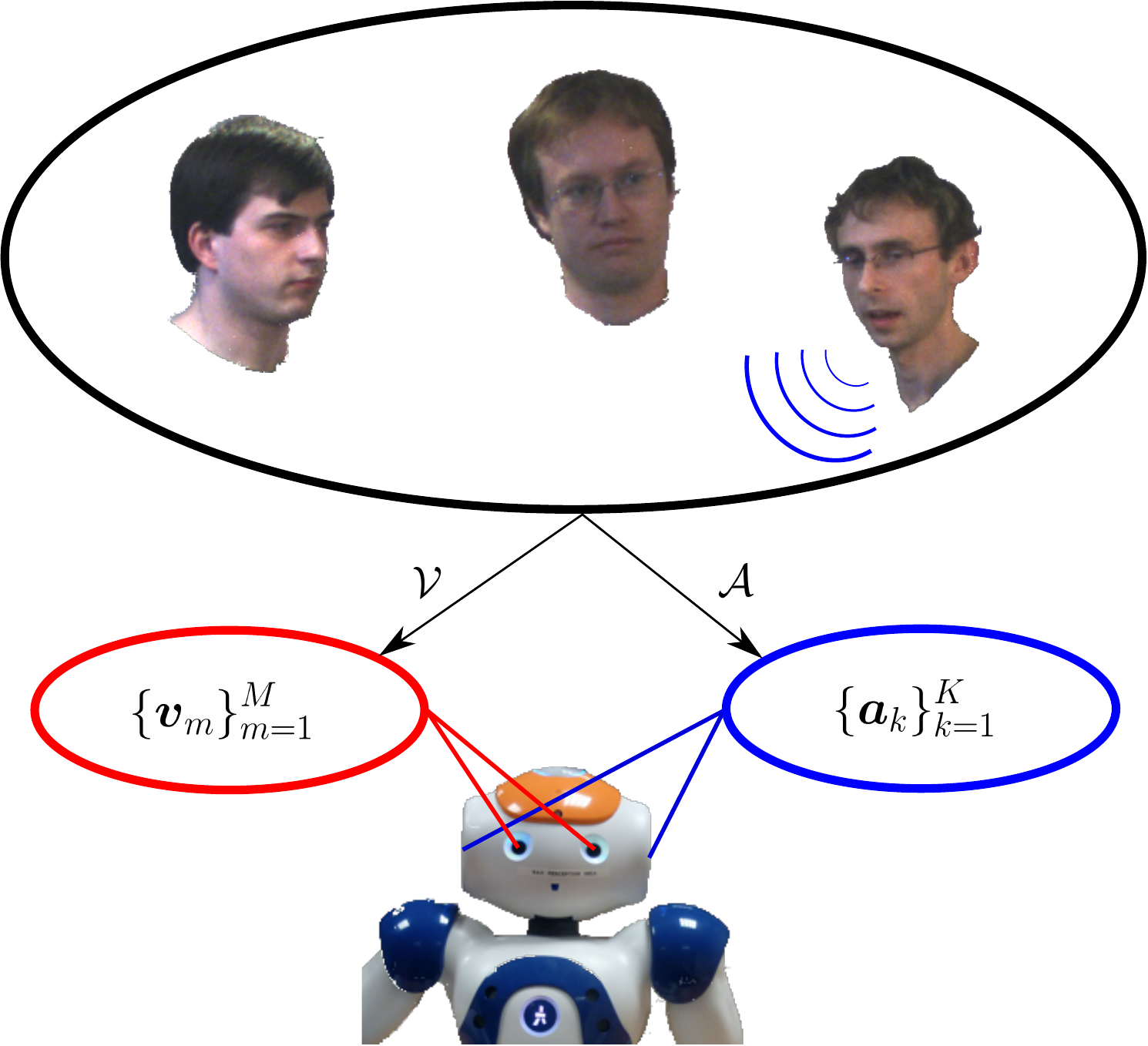}
\caption{Perceptual auditory ($\cal A$) and visual ($\cal V$) mappings of NAO. The extracted auditory
$\avect_k$ and lay around ${\cal
A}(\Svect)$ and ${\cal V}(\Svect)$ respectively. An audio-visual mapping needs to be built to link the two observations
spaces.}
\label{fig:mapping}
\end{figure}

In this section we delineate the deterministic components of our hybrid model: namely the visual and auditory mappings.
Because the scene space, the visual space and the auditory space are different we need two mappings: the first one,
${\cal A}:\mathbb{S}\rightarrow\mathbb{A}$, links the scene space to the auditory space and the second one, ${\cal
V}:\mathbb{S}\rightarrow\mathbb{V}$, links the scene space to the visual space. Both mappings are represented in
\reffig{fig:mapping}. An AV object placed at $\Svect$ in the scene space, is virtually placed at ${\cal A}(\Svect)$ in
the auditory space and at ${\cal V}(\Svect)$ in the visual space.

The definition of ${\cal A}$ and ${\cal V}$ provide a link between the two observations spaces, which corresponds
either to ${\cal A}\circ{\cal V}^{-1}$ or to ${\cal V}\circ{\cal A}^{-1}$. Depending on the extracted features and on
the sensors, the mappings ${\cal A}$ and ${\cal V}$ may be invertible. If that is not the case, ${\cal A}\circ{\cal
V}^{-1}$ or ${\cal V}\circ{\cal A}^{-1}$ should be estimated through a learning procedure. There are several works
already published dealing with this problem in different ways. In \cite{Alameda11,Sanchez12}, ${\cal V}$ is invertible
and ${\cal A}$ is known, so building ${\cal A}\circ{\cal V}^{-1}$ is straightforward. In sound source localization
approaches (inter alia \cite{Nakadai04}) ${\cal A}$ is invertible and ${\cal V}$ is known so ${\cal V}\circ{\cal
A}^{-1}$ is easily constructed. In \cite{Khalidov08,Khalidov11a}, none of the mappings are inverted, but used to tie the
parameters of the probabilistic model. So the link between $\mathbb{A}$ and $\mathbb{V}$ is not used explicitly, but
implicitly. In \cite{Butz05,Kidron05,Kidron07,Liu08}, the scene space is undetermined and the authors learn a common
representation space (the scene space) at the same time they learn both mappings.

In our case, we chose to extract 3D visual feature points, and represent them in the scene coordinate system (see
\refsec{sec:visual_features}). Thus, the mapping ${\cal V}$ is the identity, which is invertible. The auditory features
correspond to the Interaural Time Differences (see \refsec{sec:auditory_features}), and a direct path propagation model
defines ${\cal A}$. The mapping ${\cal A}\circ{\cal V}^{-1}$ is accurately built from the geometric and physical models
estimated through a calibration step (see \refsec{sec:calibration}). Consequently, we are able to map the visual
features $\vmat=\{\vvect_1,\ldots,\vvect_M\}$ onto the auditory space $\mathbb{A}$. We will denote the projection of
$\vvect_m$ by $\tilde{\vvect}_m$:
\[\tilde{\vvect}_m=({\cal A}\circ{\cal V}^{-1})(\vvect_m).\]

Summarizing, we use the mapping from $\mathbb{V}$ to $\mathbb{A}$ to map all visual features onto the auditory space.
Hence, all extracted features lie, now, in the same space, and we can perform the multimodal fusion in there.

\subsection{The Probabilistic Model}
Thanks to the link built in the previous section, we obtain a set of projected visual features
$\tilde{\vmat}=\{\tilde{\vvect}_1,\ldots,\tilde{\vvect}_M\}$, laying
in the same space as the auditory features $\amat$. These features need to be grouped to construct audio-visual
objects. However, we do not know which observation is generated by which object. Therefore, we introduce two sets of
hidden variables $\Zvect$ and $\Wvect$:
\begin{eqnarray*}
\Zvect &=& \{Z_1,\ldots,Z_m,\ldots,Z_M\} \\
\Wvect &=& \{W_1,\ldots,W_k,\ldots,W_K\},
\end{eqnarray*}
accounting for the observation-to-object assignment. The notation $Z_m=n$ ($m\in\{1,\ldots,M\}$,
$n\in\{1,\ldots,N+1\}$) means that the projected visual observation $\tilde{\vvect}_m$ was either generated by the
$n\th$ 3D object ($n\in\{1,\ldots,N\}$) or it is an outlier ($n=N+1$). Similarly, the variable $W_k$ is associated to
the auditory observation $\avect_k$.

We formulate the multimodal probabilistic fusion model under the assumption that all observations $\tilde{\vvect}_m$ and
$\avect_k$ are independent and identically distributed. The $n\th$ AV object generates both visual and auditory
features normally distributed around  ${\cal A}(\Svect_n)$ and both the visual and auditory outliers are uniformly
distributed in $\mathbb{A}$. Therefore, we write:
\begin{equation*}
\textrm{P}(\tilde{\vvect}_m|Z_m=n,\Theta) = \left\{\begin{array}{ll}
{\cal N}(\tilde{\vvect}_m;\mu_n,\sigma_n) & n=1,\ldots,N \\
{\cal U}(\tilde{\vvect}_m;\mathbb{A}) & n=N+1 \\         
\end{array}\right..
\end{equation*}
where $\Theta$ contains the Gaussian parameters, that is $\mu_n = {\cal A}(\Svect_n)$ and $\sigma_n$ (the mean and the
standard deviation of the $n\th$ Gaussian). The exact same rule holds for $\textrm{P}(\avect_m|W_m=n,\Theta)$. Thus we
can define a generative model for the observations $x\in\mathbb{A}$:
\begin{equation}
\p\left(x;\Theta\right) = \sum_{n=1}^N \pi_n\,{\cal N}(x;\mu_n,\sigma_n)
+ \pi_{N+1}\,{\cal U}(x;\mathbb{A}),
\label{eq:prob_model}
\end{equation}
where $\pi_n$ is the prior probabilities of the $n\th$ mixture component. That is,
$\pi_n=\textrm{P}(Z_m=n)=\textrm{P}(W_k=n)$, $\forall n,m,k$. The prior probabilities satisfy $\sum_{n=1}^{N+1}
\pi_n= 1$. Summarizing, the model parameters are:
\begin{equation} 
\Theta = \{\pi_1,\ldots,\pi_{N+1},\mu_1,\ldots,\mu_N,\sigma_1,\ldots,\sigma_N\}.
\end{equation}

Under the probabilistic framework described, the set of parameters is estimated within a maximum likelihood
formulation:
\begin{equation}
{\cal L}\left(\tilde{\vmat},\amat;\Theta\right) =
\sum_{m=1}^M \log\textrm{p}\left(\tilde{\vvect}_m;\Theta\right) +
\sum_{k=1}^K\log\textrm{p}\left(\avect_k;\Theta\right).
\label{eq:log-like}
\end{equation}
In other words, the optimal set of parameters is the one maximizing the log-likelihood function \refeq{eq:log-like},
where $\p$ is the generative probabilistic model in \refeq{eq:prob_model}. Unfortunately, direct maximization of
\refeq{eq:log-like} is an intractable problem. Equivalently, the expected complete-data log-likelihood will be
maximized~\cite{Dempster77} (see \refsec{sec:inference}).

We recall that the ultimate goal is to determine the number $N$ of AV events, their 3D
locations $\Svect_1,\ldots,\Svect_n,\ldots,\Svect_N$ as well as their auditory activity $e_1,\ldots,e_n,\ldots,e_N$.
However, the 3D location parameters can be computed only indirectly, once the multimodal mixture's parameters $\Theta$
have been estimated. Indeed, once the auditory and visual observations are grouped in $\mathbb{A}$, the
$\tilde{\vvect}_m \leftrightarrow \vvect_m$ correspondences are used to infer the locations $\Svect_n$ of the AV
objects and the grouping of the auditory observations $\amat$ is used to infer the speaking state $e_n$ of the AV
objects. The choice of $N$ as well as the formulas for $\Svect_n$ and $e_n$ are given in Sections~\ref{sec:bic}
and~\ref{sec:output_estimators} respectively. Before these details are given and in order to fix ideas, we devote next
section to describe the auditory and visual features, justify the existence of ${\cal V}^{-1}$ and detail the
calibration procedure leading to a highly accurate mapping ${\cal A}\circ{\cal V}^{-1}$.

\section{Finding Auditory and Visual Features}
\label{sec:features}
In this section we describe the auditory (\refsec{sec:auditory_features}) and the visual (\refsec{sec:visual_features})
features we extract from the raw data. Given this features, the definition of ${\cal A}$ and ${\cal V}$ is
straightforward. However, the computation of the mapping's parameters is done through a calibration
procedure detailed in \refsec{sec:calibration}.

\subsection{Auditory Features}
\label{sec:auditory_features}
An auditory observation $\avect_k$ corresponds to an Interaural Time Difference (ITD) between the left and right
microphones. Because the ITDs are real-valued, the auditory feature space is $\mathbb{A}=\mathbb{R}$. One ITD value
corresponds to the different of time of arrival of the sound signal between the left and right microphones. For
instance, the sound wave of a speaker located in the left-half of the scene will obviously arrive earlier to the left
microphone than to the right microphone. We found that the method proposed in \cite{Christensen07} yields
very good results that are stable over time. The relationship between an auditory source located at $\Svect
\in\mathbb{R}^3$ and an ITD observation $\avect$ depends on the relative position of the acoustic source with respect to
the locations of the left and right microphones, $\Mvect_L$ and $\Mvect_R$. If we assume direct sound propagation and
constant sound velocity $\nu$, this relationship is given by the mapping ${\cal A}:\mathbb{S}\rightarrow\mathbb{A}$
defined as:
\begin{equation}
{\cal A}(\Svect) = \frac{\|\Svect-\Mvect_L\|-\|\Svect-\Mvect_R\|}{\nu} .
\label{eq:itd}
\end{equation}

\subsection{Visual Features}
\label{sec:visual_features}
The visual observations are 3D points extracted using binocular vision. We used two types of features: the
Harris-Motion 3D (HM3D) points and the faces 3D (F3D).
\begin{description}
 \item [HM3D] The first kind of features we extracted are called Harris-motion points. We first detect Harris interest
points \cite{Harris88} in the left and right image pairs of the time interval $\Delta t$. Second, we only
consider a subset of theses points, namely those points where motion occurs. For each interest-point image location
$(u,v)$ we consider the image intensities at the same location $(u,v)$ in the subsequent images and we compute a
temporal intensity standard deviation $\tau_{(u,v)}$ for each interest point. Assuming stable lighting condition
over $\Delta t$, we simply classify the interest points into static ($\tau_{(u,v)} \leq \tau_M$) and dynamic
($\tau_{(u,v)} > \tau_M$) where $\tau_M$ is a user-defined threshold. Third, we apply a standard stereo matching
algorithm and a stereo reconstruction algorithm \cite{Hartley04} to yield a set of 3D features $\vmat$ associated with
$\Delta t$. 
\item [F3D] The second kind of features are the 3D coordinates of the speakers' faces. They are obtained
using the face detector in~\cite{Waldboost-CVPR-2005}. More precisely, the center of the bounding box retrieved by the
face detector is matched to the right image and the same stereo reconstruction algorithm as in HM3D is used to
obtain $\vmat$.
\end{description}
Both 3D features are expressed in cyclopean coordinates~\cite{Hansard08}, which are also the scene coordinates.
Consequently, the visual mapping ${\cal V}$ is the identity mapping. In conclusion, because we are able to
accurately model the geometry of the visual sensors, we can assume that ${\cal V}$ is invertible and explicitly
build the linking mapping ${\cal A}\circ{\cal V}^{-1}$.

\subsection{Calibration}
\label{sec:calibration}
\begin{figure}[t]
\centering
\includegraphics[width=0.46\textwidth]{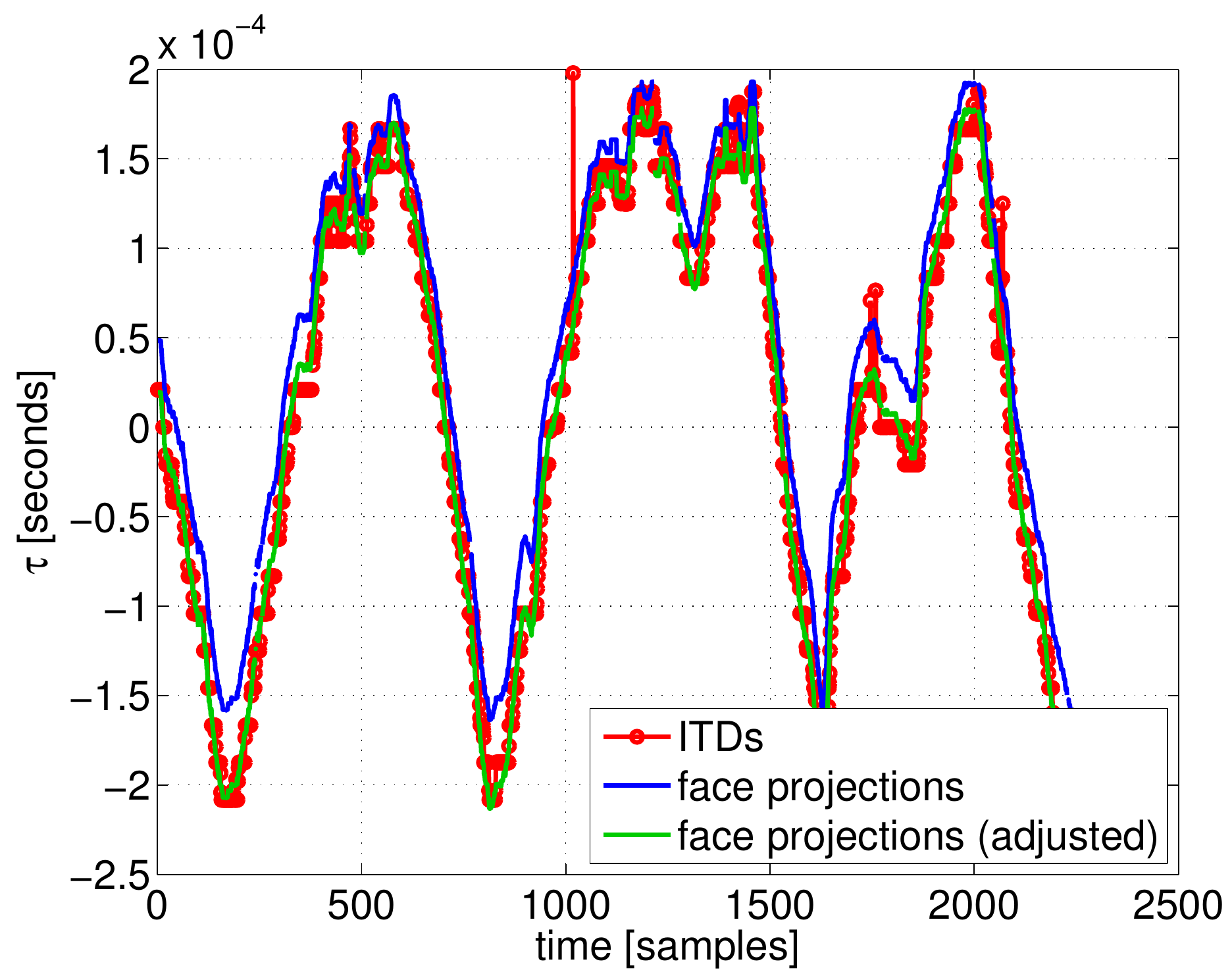}
\caption{Affine correction of the audio-visual calibration. Extracted ITD values are plot in red-circled. F3D features
projected into the ITD space using Equation \refeq{eq:itd} are plot in blue. F3D features projected using Equation
\refeq{eq:final_av_mapping}, that is after the audio-visual calibration step, are plot in green.}
\label{fig:calibration}
\end{figure}

In the two previous sections we described the auditory and the visual features respectively. As a consequence, the
mappings ${\cal A}$ and ${\cal V}$ are defined. However, we made implicit use of two, a priory unknown, objects. On one
hand the stereo-matching and the 3D reconstruction algorithms need the so-called stereo-calibration. That is, the
projection matrices corresponding to the left and right cameras which are estimated using \cite{Bouguet08}. It
is worth to remark that the calibration procedure allows us to accurately represent any point in the field-of-view of
both cameras as a 3D point. On the other hand, and in order to use ${\cal A}$, we need to know the positions of the
microphones $\Mvect_L$ and $\Mvect_R$ in the scene coordinate frame, which is slightly more complex. Since the scene
coordinates are the same as the visual coordinates, we refer to this as ``audio-visual calibration''. We manually
measure the values of $\Mvect_L$ and $\Mvect_R$ with respect to the stereo rig. However, because these measurements are
imprecise, an affine correction model needs to be applied:
\begin{equation}
\overline{\cal A}(\Svect)=c_1\,{\cal A}(\Svect)+c_0 = c_1\,\frac{\|\Svect-\Mvect_L\|-\|\Svect-\Mvect_R\|}{\nu} + c_0,
\label{eq:itd_up}
\end{equation}
where $c_1$ and $c_0$ are the adjustment coefficients. In order to estimate $c_1$ and $c_0$, a person with a
speaker held just below the face moves in a zig-zag trajectory in the entire visual field of view of the
two cameras. The 3D position of the person's face and the ITD values were extracted. We used white
noise because it correlates very well resulting in a single sharp peak in ITD space.  In many experiments, we
did not observe any effect of reverberations, because the reverberant components are suppressed by the direct component
of the long lasting white noise signal. The optimal values for $c_1$ and $c_0$, in the least square sense, were computed
from these data. \reffig{fig:calibration} shows the extracted ITDs (red-circle), the projected faces before (blue) and
after (green) the affine correction. We can clearly see how the affine transformation enhanced the audio-visual
linking mapping. Hence the projected visual features have the following expression:
\begin{equation}
\tilde{\vvect}_m = (\overline{\cal A}\circ{\cal V}^{-1})(\vvect_m) =
c_1\,\frac{\|\vvect_m-\Mvect_L\|-\|\vvect_m-\Mvect_R\|}{\nu} + c_0.
\label{eq:final_av_mapping}
\end{equation}

The outlined calibration procedure has three main advantages: (i) it requires very few training data, (ii) it lasts a
long period of time and (iii) it is environment-independent, thus guaranteeing the system's adaptability. Indeed, in our
case, the calibration ran on a one-minute audio-visual sequence and has been successfully used for the last 18
months in several rooms, including project demonstrations and conference exhibits. Consequently, the robustness of the
once-for-all tiny audio-visual calibration step is proved up to a large extent.

\section{Multimodal Inference}
\label{sec:inference}
In \refsec{sec:problem} we set up the maximum-likelihood framework to perform AV fusion. The 3D visual features are
mapped into the auditory space $\mathbb{A}$ through the audio-visual mapping $(\overline{\cal A}\circ{\cal V}^{-1})$.
This mapping takes the form in \refeq{eq:final_av_mapping} when using the auditory and visual features described in
\refsec{sec:features}. However, three of the initial issues remain unsolved: (i) the relative importance of each
modality, (ii) the estimates for $\Svect_n$ and $e_n$ and (iii) the variable number of AV objects, $N$. In this
Section we described EM-based method solving the ML problem with hidden variables and accounting for these unsolved
issues.

\subsection{Visual guidance}
\label{sec:visem}
Previous papers do not agree on how to balance the relative importance of each modality. After a deep analysis of the
features' statistics, we choose to use the visual information to guide the clustering process of the sparse auditory
observations. Indeed, because the HM3D visual features are more dense and have better temporal continuity than the ITD
values, we start by fitting a 1D GMM to the projected visual features $\{\tilde{\vvect}_m\}_{m=1}^M$. This is done with
the standard EM algorithm \cite{Bishop06}. In the E step of the algorithm the posterior probabilities $\alpha_{mn} =
\textrm{P}(Z_m=n|\tilde{\vmat},\Theta)$ are updated via the following formula:
\begin{equation}
\label{eq:alpha_m}
\alpha_{mn} = \frac{\pi_n\, \textrm{P}(\tilde{\vvect}_m|Z_m=n,\Theta)}{\sum_{i=1}^{N+1} \pi_i\,
\textrm{P}(\tilde{\vvect}_m|Z_m=i,\Theta)}.
\end{equation}

The M step is devoted to maximize the expected complete data log-likelihood with respect to the parameters, leading to
the standard formulas (with $\bar{\alpha}_n = \sum_{m=1}^M \alpha_{mn}$):
\begin{eqnarray*}
\pi_n &=& \frac{\bar{\alpha}_n}{M},  \\
 \mu_n &=& \frac{1}{\bar{\alpha}_n}\sum_{m=1}^M \alpha_{mn}\tilde{\vvect}_m, \\
\sigma_n^2 &=& \frac{1}{\bar{\alpha}_n}\sum_{m=1}^M \alpha_{mn}(\tilde{\vvect}_m-\mu_n)^2.
\end{eqnarray*}

Once the model is fitted to the projected visual data, i.e., the visual information has already been probabilistically
assigned to the $N$ objects, the clustering process proceeds by including the auditory information. Hence, we are faced
with a constrained maximum-likelihood estimation problem: maximize \refeq{eq:log-like} subject to the constraint that
the posterior probabilities $\alpha_{mn}$ were previously computed. This leads to  \textit{vision-guided EM fusion
algorithm} in which the E-step only updates the posterior probabilities associated with the auditory observations
while those associated with the visual observations remain unchanged. This semi-supervision strategy was introduced in
the context of text classification \cite{Nigam00,Miller03}. Here it is applied to enforce the quality and reliability of
one of the sensing modalities within a multimodal clustering algorithm. To summarize, the E-step of the algorithm
updates only the posterior probabilities of the auditory observations $\beta_{kn}=\textrm{P}(W_k =
n|\amat,\Theta)$:
\begin{equation}
\label{eq:beta_kn}
\beta_{kn} =  \frac{\pi_n\,\textrm{P}(\avect_k|W_k = n,\Theta)}{\sum_{i=1}^{N+1}\pi_i\,\textrm{P}(\avect_k|W_k =
i,\Theta)},
\end{equation}
while keeping the visual posterior probabilities, $\alpha_{mn}$, constant. 
The M-step has a closed-form solution and the prior probabilities are updated with:
\[ \pi_n = \frac{\gamma_n}{M+K},\quad n=1,\ldots,N+1, \]
with $\gamma_n = \sum_{m=1}^M \alpha_{mn} + \sum_{k=1}^K \beta_{kn} = \bar{\alpha}_n + \bar{\beta}_n$.
The means and variances of the current model are estimated by combining the two modalities:
\begin{equation}
\mu_n = \frac{1}{\gamma_n} \left(\sum_{m=1}^M \alpha_{mn}\,\tilde{\vvect}_m + \sum_{k=1}^K
\beta_{kn}\,\avect_k\right),
\label{eq:update_mu}
\end{equation}
\begin{equation}
\sigma^2_n = \frac{\sum_{m=1}^M \alpha_{mn}\,(\tilde{\vvect}_m-\mu_n)^2 +
\sum_{k=1}^K \beta_{kn}\,(\avect_k-\mu_n)^2}{\gamma_n}.
\label{eq:update_sigma}
\end{equation}

\subsection{Counting the number of speakers}
\label{sec:bic}
Since we do not know the value of $N$, a reasonable way to proceed is to estimate the parameters $\Theta_N$ for
different values of $N$ using the method delineated in the previous section. Once we estimated the maximum likelihood
parameters for models with different number of AV objects, we need a criterion to choose which is the best one. This is
estimating the number of AV objects (clusters) in the scene. BIC~\cite{Schwarz78} is a well known criterion to choose
among several maximum likelihood statistical models. BIC is often chosen for this type of tasks due to its attractive
consistency properties~\cite{Keribin00}. It is appropriate to use this criterion in our framework, due to the fact that
the statistical models after the \textit{vision-guided EM algorithm}, fit the AV data in an ML sense. In our case,
choosing among these models is equivalent to estimate the number of AV events $\hat{N}$. The formula to compute
the BIC score is:
\begin{equation}
\textrm{BIC}(\tilde{\vvect},\amat,\Theta_N) = {\cal L}\left(\tilde{\vvect},\amat;\Theta_N\right) -
\frac{D_N\,\log(M+K)}{2},
\label{eq:criteria}
\end{equation}
where $D_N = 3N$ is the number of free parameters of the model.

The number of AV events is estimated by selecting the statistical model corresponding to the maximum score:
\begin{equation}
\hat{N} = \arg\max\limits_N \textrm{BIC}(\tilde{\vvect},\amat,\Theta_N).
\label{eq:estimate-N}
\end{equation}

\subsection{Detection and localisation}
\label{sec:output_estimators}
The selection on $N$ leads to the best maximum-likelihood model in the BIC sense. That is, the set of parameters that
best explain the auditory and visual observations $\amat$ and $\tilde{\vmat}$. In the following, $\vmat$ are used
to estimate the 3D positions in the scene and $\amat$ to estimate the speaking state of each AV object. 

The locations of the AV objects are estimated thanks to the one-to-one correspondence between 3D visual features and
the 1D projected features, $\tilde{\vvect}_m\leftrightarrow\vvect_m$. Indeed, the probabilistic assignments of the
projected visual data onto the 1D clusters, $\alpha_{mn}$, allow us to estimate $\Svect_n$ through:
\begin{equation}
\hat{\Svect}_n = \frac{1}{\bar{\alpha}_n} \sum_{m=1}^M \alpha_{mn} \vvect_m.
\label{eq:hat_svect}
\end{equation}

The auditory activity associated to the $n\th$ speaker is estimated as follows ($\tau_A$ is a user-defined threshold):
\begin{equation} 
\hat{e}_n = 
\left\{
\begin{array}{ll} 
1& \mbox{if } \bar{\beta}_n > \tau_A\\
 0 & \mbox{otherwise}
\end{array}
\right. 
\label{eq:audio_act}
\end{equation}

This two formulas account for the last remaining issue: the 3D localization and speaking state estimation of the AV
objects. Next section describes some practical considerations to take into account when using this EM-based AV fusion
method. Afterward, in \refsec{sec:mgrh}, we summarize the method by providing an algorithmic scheme of the
multimodal inference procedure.

\subsection{Practical Concerns}
\label{sec:practical_concerns}
Even though the EM algorithm has proved to be the proper (and extremely powerful) methodology to solve ML problems with
hidden variables, in practice we need to overcome two main hurdles. First, since the log-likelihood function has many
local maxima and EM is a local optimization technique, a very good initialization is required. Second, because real
data is finite and may not strictly follow the generative law of probability \refeq{eq:prob_model}, the consistency
properties of the EM algorithm do not guarantee that the model chosen by BIC is meaningful regarding the application.
Thus, a post-processing step is needed in order to include the application-dependant knowledge. In all, we must
account for three practical concerns: (i) EM initialization, (ii) eventual shortage of observations and (iii) the
probabilistic model does not fully correspond to the observations.

It is reasonable to assume that the dynamics of the AV objects are somehow constrained. In other words, the positions
of the objects at a time interval are close to the positions at the previous time interval. Hence, we use the model
computed in the previous time interval to initialize the EM based procedure. More precisely, if we denote by $N^{(p)}$
the number of AV objects found in the previous time interval, we initialize a new 1D GMM with $N$ clusters,
for $N\in\{0,\ldots,N_{\textrm{max}}\}$. In the case $N\leq N^{(p)}$, we take the $N$ clusters with the
highest weight. For $N>N^{(p)}$, we incrementally split a cluster at its mean into two clusters. The  cluster to be
split is selected on the basis of a high Davies-Bouldin index \cite{Davies79}: 
\[ DW_i = \max_{j\neq i} \frac{\sigma_i + \sigma_j}{\|\mu_i - \mu_j\|}. \]

We chose to split the cluster into two clusters in order to detect AV objects that have recently appeared in
the scene, either because they were outside the field of view, or because they were occluded by another AV object. This
provides us with a good initialization. In our case the maximum number of AV objects is $N_{\textrm{max}} = 10$.

A shortage of observations usually leads to clusters whose interactions may describe an overall pattern, instead of
different components. We solve this problem by merging some of the mixture's components. There are several techniques to
merge clusters within a mixture model (see \cite{Henning10}). Since the components to be merged lie around the same
position and have similar spread, the \textit{ridgeline} method \cite{Surajit05} best solves our problem.

Finally, we need to face the fact that the probabilistic model does not fully represent the observations. Indeed, we
observed the existence of spurious clusters. Although the 3D visual observations associated with these clusters may be
uniformly distributed, their projections onto the auditory space $\tilde{\vvect}_m$ may form a spurious cluster. Hence
these clusters are characterized by having their points distributed near some hyperboloid in the 3D space (hyperboloids
are the level surfaces of the linking mapping defined in \refeq{eq:final_av_mapping}). As a
consequence, the volume of the back-projected 3D cluster is small. We discard those clusters whose covariance matrix
has a small determinant. Similarly as in \refeq{eq:hat_svect}, the clusters' covariance matrix is estimated
via:
\begin{equation}
\hat{\Sigmamat}_n = \frac{1}{\bar{\alpha}_n} \sum_{m=1}^M \alpha_{mn}
\left(\vvect_m-\hat{\Svect}_n\right)\left(\vvect_m-\hat{\Svect}_n\right)^\top.
\label{eq:hat_svect}
\end{equation}

\subsection{Motion-Guided Robot Hearing}
\label{sec:mgrh}
\refalg{alg:procedure} below summarizes the proposed method. It takes as input the visual (MH3D) and
auditory (ITD) observations gathered during a time interval $\Delta t$. The algorithm's output is the estimated
number of clusters $\hat{N}$, the estimated 3D positions of the AV events $\{\hat{\Svect}_n\}_{n=1}^{\hat{N}}$ as well
as their estimated auditory activity $\{\hat{e}_n\}_{n=1}^{\hat{N}}$. Because the grouping process is supervised by the
HM3D features, we name the procedure \textit{Motion-Guided Robot Hearing}. The algorithm starts by mapping the visual
observations onto the auditory space by means of the linking mapping defined in \refeq{eq:final_av_mapping}. Then, for
$N\in\{1,\ldots,N_\textrm{max}\}$ it iterates through the following steps: (a)~Initialize a model with $N$ components
using the output of the previous time interval (\refsec{sec:practical_concerns}), (b) apply EM using the selected $N$ to
model the 1D projections of the visual data (\refsec{sec:visem}), (c) apply the \textit{vision-guided EM fusion}
algorithm to both the auditory and projected visual data (\refsec{sec:visem}) in order to perform audio-visual
clustering, and (d) compute the BIC score associated with the current model, i.e., \refeq{eq:criteria}. This allows the
algorithm to select the model with  the highest BIC score, i.e., \refeq{eq:estimate-N}. The post-processing step is then
applied to the selected model (\refsec{sec:practical_concerns}) prior to computing the final output
(\refsec{sec:output_estimators}).

\begin{algorithm}
\caption{Motion-Guided Robot Hearing}
\label{alg:procedure}
\begin{algorithmic}[1]
\STATE \textbf{Input:} HM3D, $\left\{\vvect_m\right\}_{m=1}^M$, and ITD, $\left\{\avect_k\right\}_{k=1}^K$, features.
\STATE \textbf{Output:} Number of AV events $\hat{N}$, 3D localization $\left\{\hat{\Svect}_n\right\}_{n=1}^{\hat{N}}$
and auditory status $\left\{\hat{e}_n\right\}_{n=1}^{\hat{N}}$.
\STATE Map the visual features onto the auditory space, $\tilde{\vvect}_m = (\overline{\cal A}\circ{\cal
V}^{-1})(\vvect_m)$ \refeq{eq:final_av_mapping}.
\FOR{$N = 1 \to N_\textrm{max}$}
\STATE \textbf{(a) } { Initialize the model with $N$ clusters (\refsec{sec:practical_concerns}).}
\STATE \textbf{(b) } { Apply EM clustering to $\{\tilde{\vvect}_m\}_{m=1}^M$ (\refsec{sec:visem}).}
\STATE \textbf{(c) } { Apply the \textit{Vision-guided EM fusion} algorithm to cluster the audio-visual data
(\refsec{sec:visem}).}
\STATE \textbf{(d) } { Compute the BIC score \refeq{eq:criteria}}.
\ENDFOR
\STATE Estimate the number of clusters based on the BIC score \refeq{eq:estimate-N}.
\STATE Post-processing (\refsec{sec:practical_concerns}).
\STATE Compute the final outputs $\{\hat{\Svect}_n\}_{n=1}^{\hat{N}}$ and $\{\hat{e}_n\}_{n=1}^{\hat{N}}$
(\refsec{sec:output_estimators}).
\end{algorithmic}
\end{algorithm}

\section{Implementation on NAO}
\label{sec:robot_implem}
The previous multimodal inference algorithm has desirable statistical properties and good performance (see
\refsec{sec:results}). Since our final aim is to have a stable component working on a humanoid robot (i.e., able to
interact with other components), we reduced the computational load of the AV fusion algorithm. Indeed, we adapted the
method described in \refsec{sec:inference} to achieve a light on-line algorithm working on mobile robotic platforms.

In order to reduce the complexity, we substituted the Harris-Motion 3D point detector (HM3D) with the face 3D detector
(F3D), described in \refsec{sec:visual_features}. F3D replaces hundreds of HM3D points with a few face locations in 3D,
$\{\vvect_m\}_{m=1}^M$. We then consider that the potential speakers correspond to the detected faces. Hence we set
$N=M$ and $\Svect_n=\vvect_n$, $n=1,\ldots,N$. This has several crucial consequences. First, the number of AV objects
corresponds to the number of detected faces; the model selection step is not needed and the EM algorithm does
not have to run $N_{\textrm{max}}$ times, but just once. Second, because the visual features provide a good
initialization for the EM (by setting $\mu_n=(\overline{\cal A}\circ{\cal V}^{-1})(\Svect_n)$), the visual EM is not
required and the hidden variables $\Zvect$ do not make sense anymore. Third, since the visual features are not used as
observations in the EM, but to initialize it, the complexity of the \textit{vision-guide EM fusion} algorithm is ${\cal
O}(NK)$ instead of ${\cal O}\left(N(K+M)\right)$. This important because the number of HM3D points is much bigger than
the number of ITD values, i.e., $M\gg K$. Last, because the visual features provide the $\Svect_n$'s, there is not need
to estimate them through \refeq{eq:hat_svect}.

\subsection{Face-Guided Robot Hearing}
\label{sec:fgrh}
The resulting procedure is called \textit{Face-Guided Robot Hearing} and it is summarized
in \refalg{alg:procedure_simple} below. It takes as input the detected heads ($\Svect_1,\ldots,\Svect_N$) and the
auditory ($\amat$) observations gathered during a time interval $\Delta t$. The algorithm's output is the estimated
auditory activity $\{\hat{e}_n\}_{n=1}^{N}$.

\begin{algorithm}
\caption{Face-Guided Robot Hearing}
\label{alg:procedure_simple}
\begin{algorithmic}[1]
\STATE \textbf{Input:} Faces' position $\{\Svect_n\}_{n=1}^N$ and auditory $\left\{\avect_k\right\}_{k=1}^K$
features.
\STATE \textbf{Output:} AV objects' auditory status $\left\{\hat{e}_n\right\}_{n=1}^{\hat{N}}$.
\STATE Map the detected heads onto the auditory space, $\mu_n = (\overline{\cal A}\circ{\cal V}^{-1})(\Svect_n)$
\refeq{eq:final_av_mapping}.
\STATE Apply EM clustering to $\{\avect_k\}_{k=1}^K$ (\refsec{sec:visem}).
\STATE Compute the final outputs $\{\hat{e}_n\}_{n=1}^{\hat{N}}$ (\refsec{sec:output_estimators}).
\end{algorithmic}
\end{algorithm}

\subsection{System Architecture}
\label{sec:architecture}
We implemented our method using several components which are connected by a middleware called
Robotics Services Bus (RSB) \cite{Wienke2011}. RSB is a platform-independent event-driven middleware specifically
designed for the needs of distributed robotic applications. It is based on a logically unified bus which can span over
several transport mechanisms like network or in-process communication. The bus is hierarchically structured using scopes
on which events can be published with a common root scope. Through the unified bus, full introspection of the event flow
between all components is easily possible. Consequently, several tools exist which can record the event flow and replay
it later, so that application development can largely be done without a running robot. RSB events are automatically
equipped with several timestamps, which provide for introspection and synchronization abilities. Because of these
reasons RSB was chosen instead of NAO's native framework NAOqi and we could implement and test our algorithm remotely
without performance and deployment restrictions imposed by the robot platform. Moreover, the resulting implementation
can be reused for other robots.

One tool available in the RSB ecosystem is an event synchronizer, which synchronizes
events based on the attached timestamps with the aim to free application developers from such a generic task. However,
several possibilities of how to synchronize events exist and need to be chosen based on the intended application
scenario. For this reason, the synchronizer implements several strategies, each of them synchronizing events from
several scopes into a resulting compound event containing a set of events from the original scopes. We used two
strategies for the implementation.
The \emph{ApproximateTime} strategy is based on the algorithm available in \cite{ApproximateTime} and outputs sets of
events containing exactly one event from each scope. The algorithm tries to minimize the time between the earliest and
the latest event in each set and hence well-suited to synchronize events which originate from the same source (in the
world) but suffered from perception or processing delays in a way that they have non-equal timestamps. The
second algorithm, \emph{TimeFrame}, declares one scope as the primary event source and for each event received here, all
events received on other scopes are attached that lie in a specific time frame around the timestamp of the source event.

\emph{ApproximateTime} is used in our case to synchronize the results from the left and right camera as frames in
general form matching entities but due to independent grabbing of both cameras have slightly different timestamps.
Results from the stereo matching process are synchronized with ITD values using the \emph{TimeFrame} strategy because
the integration time for generating ITD values is much smaller than for a vision frame and hence multiple ITD values
belong to a single vision result.

\subsection{Modular Structure}
\label{sec:modular_structure}
\begin{figure}[t]
\centering
\includegraphics[width=0.6\linewidth]{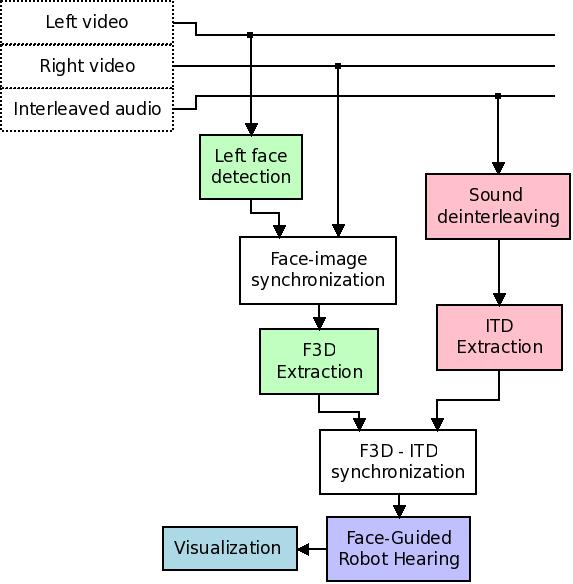}
\caption {Modular structure of the \textit{Face-Guided Robot Hearing} procedure implemented on NAO. There are five types
of modules: streaming \& synchronization (white), visual processing (green), auditory processing (red), audio-visual
fusion (purple) and visualization (blue).}
\label{fig:pipeline}
\end{figure}

The implementation is divided into components shown in the pipeline of \reffig{fig:pipeline}. Components are
color-coded: modules provided by the RSB middleware (white), auditory (red) and visual (green) processing, audio-visual
fusion (purple) and the visualization tool (blue) described at the end of this Section.

\begin{figure}
 \centering
 \includegraphics[width=0.4\linewidth]{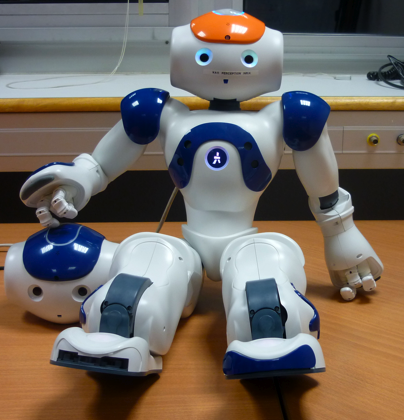}
 \caption{Within this work we used a new audio-visual head that is composed of a synchronized camera pair and two
microphones. This ``orange'' head replaces the former ``blue'' head and is fully interfaced by the RSB middleware
previously described in this section.}
 \label{fig:nao_head}
\end{figure}

The visual processing is composed by five modules. \emph{Left video} and \emph{Right video} stream the images
received at left and right cameras. The \emph{Left face detection} module extracts the faces from the left image.
These are then synchronized with the right image in \emph{Face-image synchronization}, using the \emph{ApproximateTime}
strategy. The \emph{F3D Extraction} module computes the F3D features. A new audio-visual head for NAO was used for this
implementation. The new head (see \reffig{fig:nao_head}) is equipped with a pair of cameras and four microphones, thus
providing a synchronized VGA stereoscopic image flow as well as four audio channels.

The auditory component consists of three modules. Interleaved audio samples coming from the four microphones of NAO are
streamed by the \emph{Interleaved audio} module. The four channels are deinterleaved by the \emph{Sound
deinterleaving} module, which outputs the auditory flows corresponding to the left and right microphones. These flows
are stored into two circular buffers in order to extract the ITD values (\emph{ITD extraction} module).

Both visual and auditory features flow until the \emph{Audio-visual synchronization} module; the \emph{TimeFrame}
strategy is used here to find the ITD values coming from the audio pipeline associated to the 3D positions of the
faces coming from the visual processing. These synchronized events feed the \emph{Face-guided robot hearing} module,
which is in charge of estimating the speaking state of each face, $e_n$.

\begin{figure}
\centering
\includegraphics[width=0.6\textwidth]{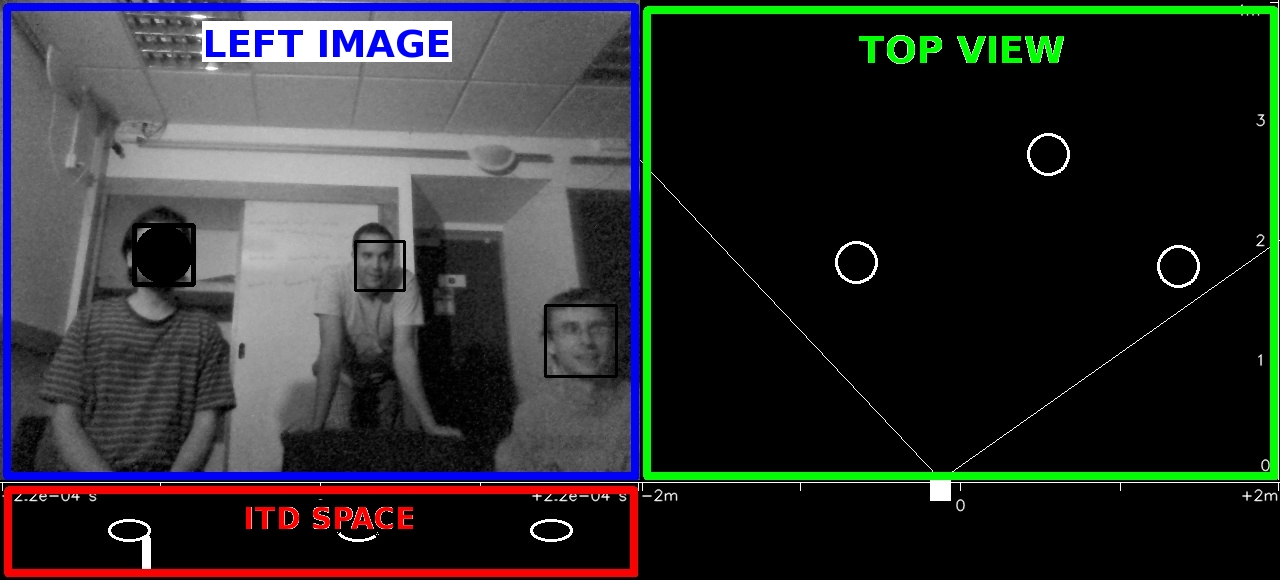}
\caption {Snapshot of the visualization tool. The top-left (blue-framed) image is the original left image plus one
bounding box per detected face. In addition, an intensity-coded circle appears when the speaker is active. The darker
the color is, the higher the speaking probability is. The top-right (green-framed) image corresponds to the bird-view of
the scene, in which each circle corresponds to a detected head. The bottom-left (red-framed) image represents the ITD
space. The projected faces are represented by an ellipse and the histogram of extracted ITD values is plot.}
\label{fig:vtool}
\end{figure}

Finally, we developed the module \emph{Visualization}, in order to get a better insight of the proposed algorithm. A
snapshot of this visualization tool can be seen in \reffig{fig:vtool}. The image consists of three parts. The top-left
part with a blue frame is the original left image plus one rectangle per detected face. In addition to the face's
bounding box, a solid circle is plot on the face of the actor codding the emitting sound probability, the higher it is,
the darker the circle. The top-right part, framed in green, is a bird-view of the scene, in which the detected heads
appear as circles. The bottom-left part, with a red frame, represents the ITD space. There, both the mapped heads
(ellipses) and the histogram of ITD values are plot.

\subsection{Implementation Details}
Some details need to be specified regarding the implementation of the face-guided robot hearing method. First, the
integration window $F$ and the frame shift $f$ of the ITD extraction procedure. The bigger the integration window is
the more reliable the ITD values are and the more expensive its computation becomes. Similarly, the smaller $f$ is the
more ITD observations are extracted and the more computational load we have. A good compromise between low computational
load, high rate, and reliability of ITD values was found for $W=150$~ms and $f=20$~ms. We also used an activity
threshold: when the energy of the sound signals is lower than $E_A=0.001$, the window is not processed. Thus saving
computational time for other components in the system when there are no emitted sounds. Notice that this parameter could
be controlled by a higher level module which would learn the characteristics of the scene and infer the level of
background noise. We initialize $\sigma_n^2=10^{-9}$, since we found this value big enough to take into account the
noise in the ITD values and small enough to discriminate speakers that are close to each other. The threshold $\tau_A$
has to take into account how many audio observations ($K$) are gathered during the current time interval $\Delta t$ as
well as the number of potential audible AV objects ($N$). For instance, if there is just one potential AV object, most
of the audio observations should be assigned to it, whereas if there are three of them the audio observations may be
distributed among them (in case all of them emit sounds). The threshold $\tau_A$ was experimentally set to $\tau_A = K /
(N+2)$. The entire pipeline was running on a laptop with an i7 processor at $2.5$~GHz. 

\section{Results}
\label{sec:results}
In order to evaluate the proposed approach, we ran three sets of experiments. First, we evaluated the Multimodal
Inference method described in \refsec{sec:inference} on synthetic data. This allowed us to assess the quality of the
model on a controlled scenario, where the feature extraction did not play any role. Second, we evaluated the
\textit{Motion-Guided Robot Hearing} method on a publicly available dataset, thus assessing the quality of the entire
approach. Finally, we evaluated the \textit{Face-Guided Robot Hearing} implemented on NAO, which proves that the
proposed hybrid deterministic/probabilistic framework is suitable for robot applications.

\begin{figure}
\centering
\begin{tabular}{cc}
 \subfloat[]{\includegraphics[width=0.3\columnwidth]{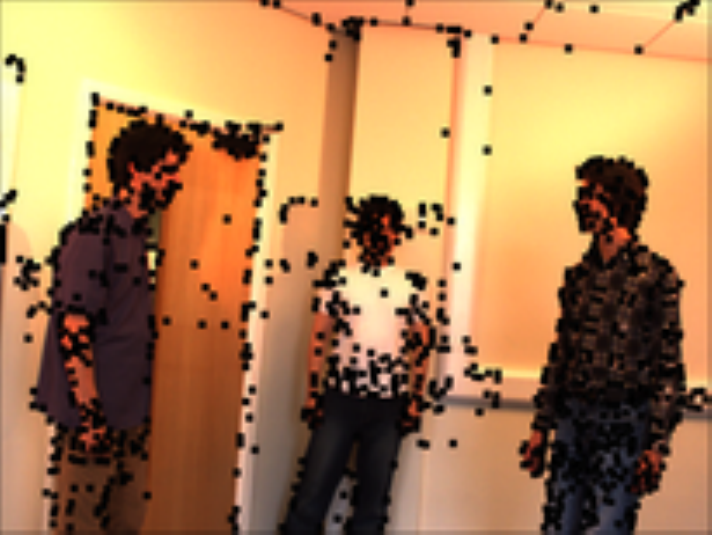}} &
 \subfloat[]{\includegraphics[width=0.3\columnwidth]{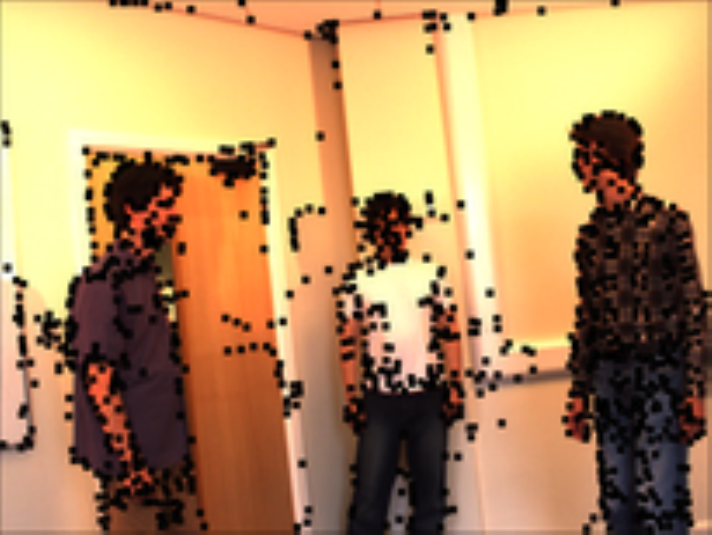}} \\
 \subfloat[]{\includegraphics[width=0.3\columnwidth]{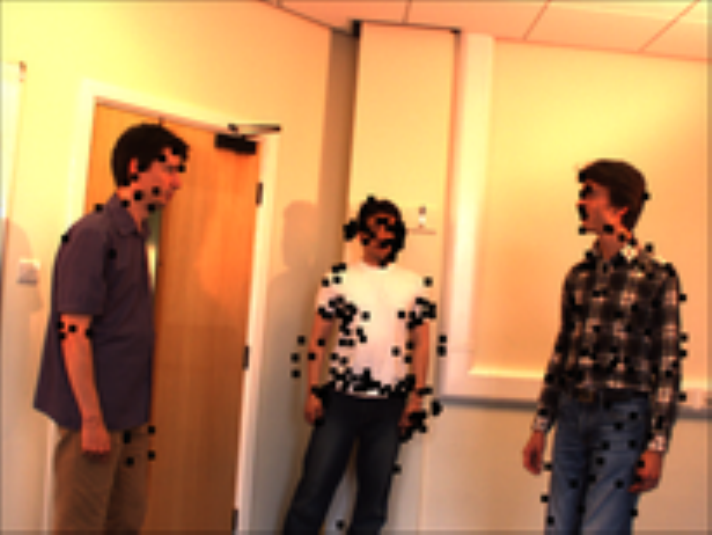}} &
 \subfloat[]{\includegraphics[width=0.3\columnwidth]{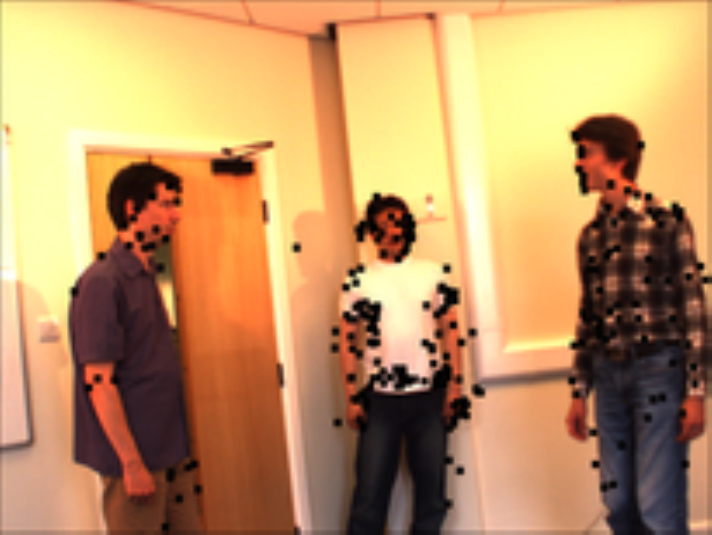}}\\
 \multicolumn{2}{c}{\subfloat[]{\includegraphics[width=0.3\columnwidth]{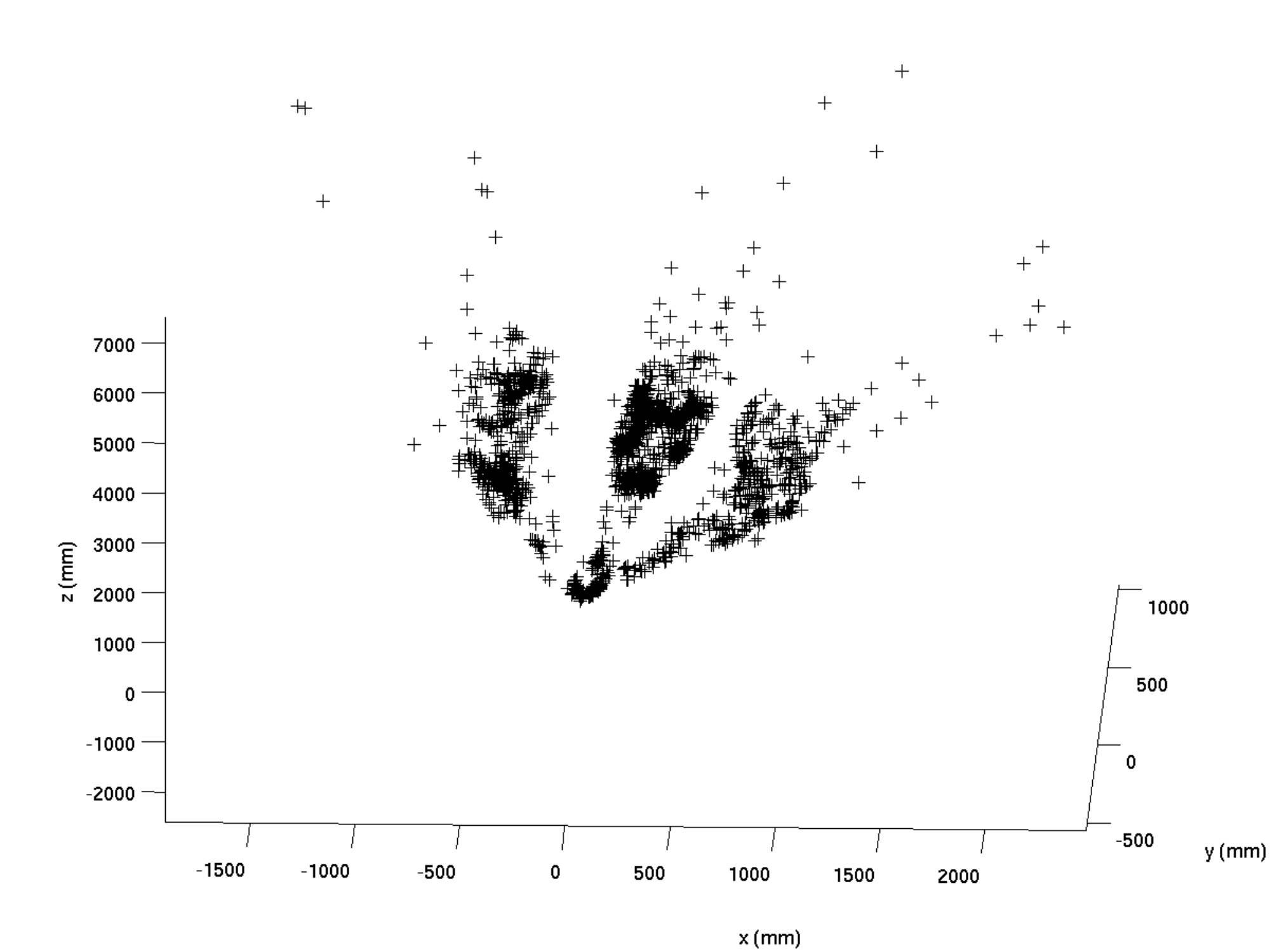}}}\\
\end{tabular}
\caption{Interest points as detected in the left (a) and right (b) images. Dynamic interest points detected in the left
(c) and the right (d) images. (e) HM3D visual observations, $\{\vvect_m\}_{m=1}^M$. Most of the background (hence
static) points are filtered out from (a) to (c) and from (b) to (d). It is worth noticing that the reconstructed HM3D
features suffer from reconstruction errors.}
\label{fig:example-Vobservations}
\end{figure}

\begin{figure}
 \centering
\begin{tabular}{cc}
 \subfloat[]{\includegraphics[width=0.3\columnwidth]{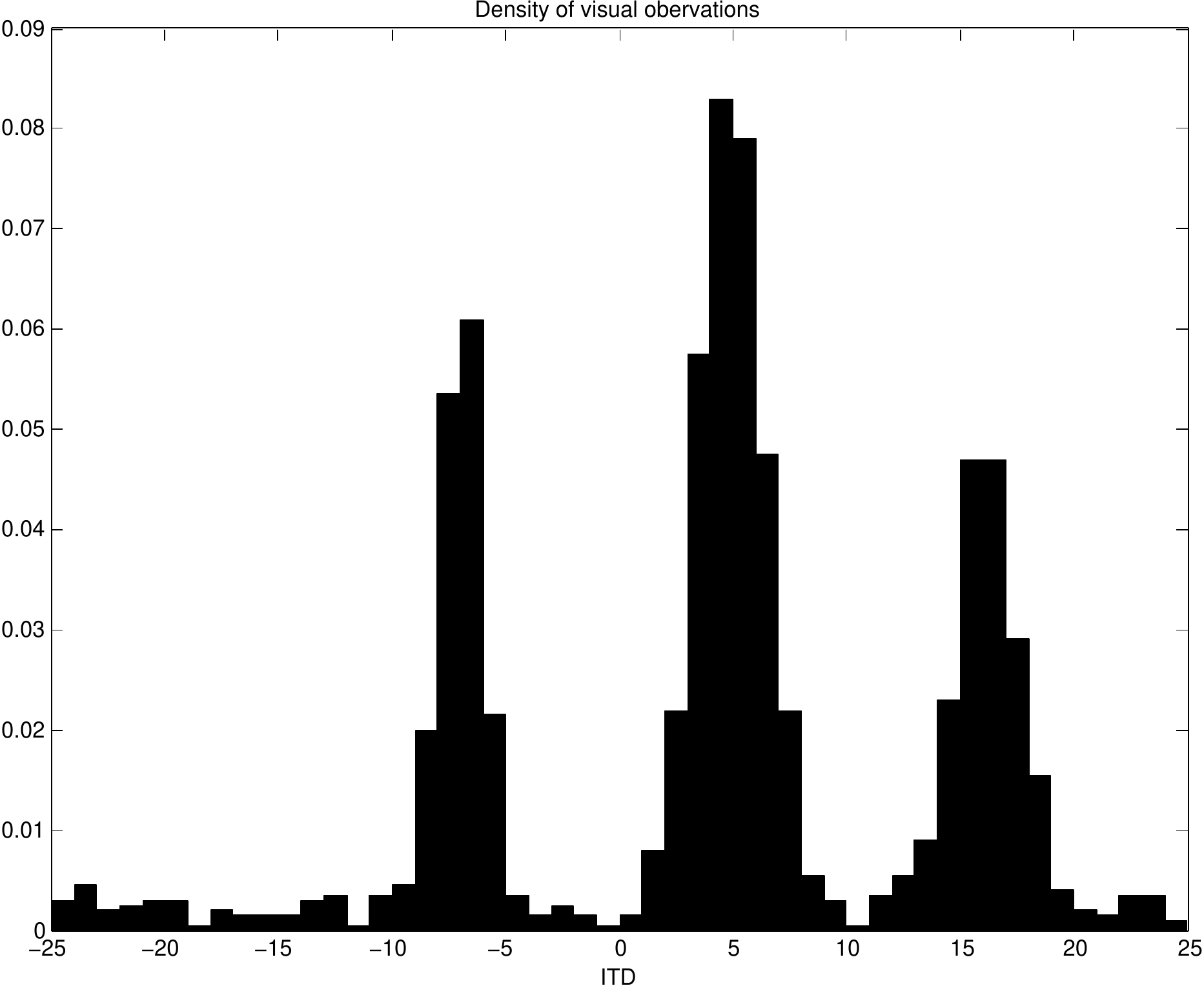}} &
 \subfloat[]{\includegraphics[width=0.3\columnwidth]{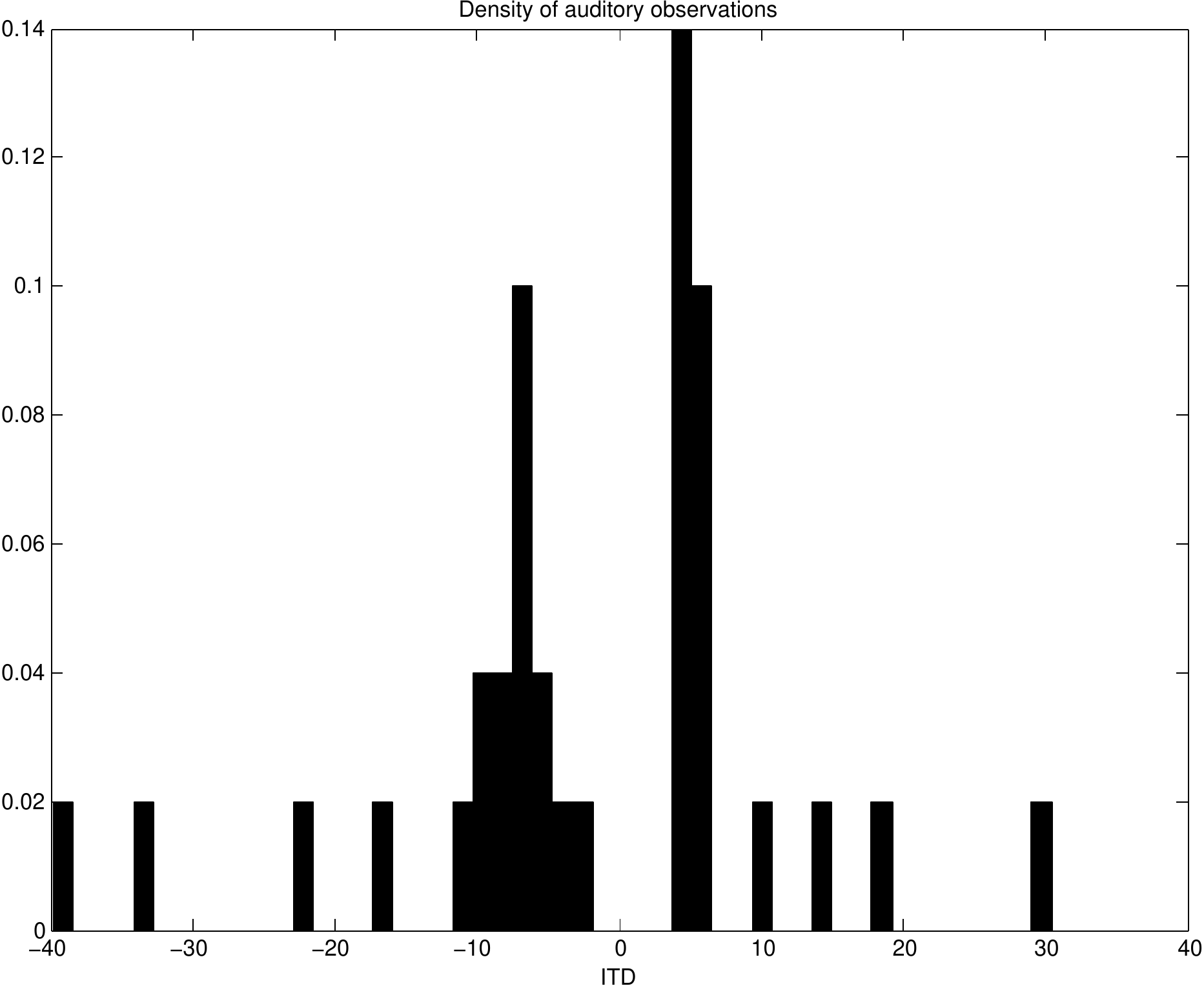}}
\end{tabular}
\caption{Observation densities in the auditory space $\mathbb{A}$: (a) of the projected HM3D features,
$\{\tilde{\vvect}_m\}_{m=1}^M$, and (b) of the ITD features, $\{\avect_k\}_{k=1}^K$. In this particular example, we
observe three moving objects (corresponding to the three people in the images). In addition, two of them are emitting
sound (left and middle) and one is silent (right). We remark that auditory as well as visual observations are
contaminated by noise (enlarging the Gaussian variances) and by outliers (uniformly distributed in the auditory
feature space).}
\label{fig:example-AVobservations}
\end{figure}

In all our experiments we used a time interval of 6 visual frames, $\Delta t=0.4s$; time in which approximately 2,000
HM3D observations and 20 auditory observations are extracted. A typical set of visual and auditory observations are
shown in Figures~\ref{fig:example-Vobservations} and~\ref{fig:example-AVobservations}. Indeed,
\reffig{fig:example-Vobservations} focuses on the extraction of the HM3D features: the Harris interest point
detection, filtered by motion, matched between images and reconstructed in 3D. \reffig{fig:example-AVobservations}
shows the very same 3D features projected in to the ITD space. Also, the ITD values extracted during the same time
interval are shown. These are the input features of the \textit{Motion-Guided Robot Hearing} procedure. Notice that both
auditory and visual data are corrupted by noise and by outliers. Visual data suffer from reconstruction errors either
from wrong matches or from noisy detection. Auditory data suffer from reverberations, which enlarge the pics' variances,
or from sensor noise which is sparse along the ITD space.

To quantitatively evaluate the localization results, we compute a distance matrix between the detected clusters and
the ground-truth clusters. The cluster-to-cluster distance corresponds to the Euclidean distance between cluster means.
Let $\mathbf{D}$ be the distance matrix, then entry $D_{ij}=\|\mu_i-\hat{\mu}_j\|$ is the distance from the $i\th$
ground-truth cluster to the $j\th$ detected cluster. Next, we associate at most one ground-truth cluster to each
detected cluster. The assignment procedure is as follows. For each detected cluster we compute its ground-truth
nearest cluster. If it is not closer than a threshold $\tau_{\textrm{loc}}$ we mark it as a \textit{false positive},
otherwise we assign the detected cluster to the ground-truth cluster. Then, for each ground-truth cluster we determine
how many detected clusters are assigned to it. If there is none, we mark the ground-truth cluster as \textit{false
negative}. Finally, for each of the remaining ground-truth clusters, we select the closest (\textit{true positive})
detected cluster among the ones assigned to the ground-truth cluster and we mark the remaining ones as \textit{false
positives}. We can evaluate the localization error and the auditory state for those clusters that have been
correctly detected . The localization error corresponds to the Euclidean distance between the means. Notice that by
choosing $\tau_{\textrm{loc}}$, we fix the maximum localization error allowed. The auditory state is counted as
\textit{false positive} if detected audible when silent, \textit{false positive} if detected silent when audible and
\textit{true positive} otherwise. $\tau_{\textrm{loc}}$ was set to $0.35\,\textrm{m}$ in all the experiments. 

\subsection{Results on Synthetic Data}
\label{sec:synthetic_data}
Four synthetic sequences containing one to three AV objects were generated. These objects can move and they are not
necessarily visible/audible along the entire sequence. \reftab{tab:video_results_synth} shows the visual evaluation of
the method when tested with synthetic sequences. The sequence code name describes the dynamic character of the sequence
(\textit{Sta} means static and \textit{Dyn} means dynamic) and the varying number of AV objects in the scene
(\textit{Con} means constant number of AV objects and \textit{Var} means varying number of AV objects). The columns show
different evaluation quantities: FP (\textit{false positives}), i.e., AV objects found that do not really exist, FN
(\textit{false negatives}), i.e., present AV objects that were not found, TP (\textit{true positives}) and ALE (average
localization error). Recall that we can compute the localization error just for the true positives. First, we observe
that the right detection rate is always above 65\%, increasing to 96\% in the case where there are 3 visible static
clusters. We also observe that the fact that the number of AV objects in the scene varies does not impact the
localization error. The effect on the localization error is due, hence, to the dynamic character of the scene; if the AV
objects move or not. The third observation is that both the dynamic character of the scene and the varying number of
clusters have a lot of impact on the detection rate.

\begin{table}
\caption{Visual evaluation of results obtained with synthetic sequences. \textit{Sta}/\textit{Dyn} states for static or
dynamic scene; the AV objects move or do not move. \textit{Var}/\textit{Con} states for varying or constant number of AV
objects. FP stands for false positives, FN for false negatives, TP for true positives and ALE for average
localization
error (expressed in meters).\vspace{0.2cm}}
\label{tab:video_results_synth}
\centering
{\small
\begin{tabular}{lcccc}
\toprule
Seq. & FP & FN & TP & ALE [m] \\
\midrule
\textit{StaCon} & 12 &  16 (3.9\%) & 392 (96.1\%) & 0.03 \\
\textit{DynCon} & 43 & 139 (34.1\%) & 269 (65.9\%) & 0.10 \\
\textit{StaVar} & 46 & 69 (30.1\%) & 160 (69.9\%) & 0.03 \\
\textit{DynVar} & 40 & 82 (35.9\%) & 147 (64.1\%) & 0.11 \\
\bottomrule
\end{tabular}}
\end{table}

\begin{table}
\caption{Audio evaluation of the results obtained with synthetic sequences. \textit{Sta}/\textit{Dyn} states for static
or dynamic scene; the AV objects move or do not move. \textit{Var}/\textit{Con} states for varying or constant number of
AV objects.\vspace{0.2cm}}
\label{tab:audio_results_synth}
\centering
{\small
\begin{tabular}{lccc}
\toprule
Seq. & FP & FN & TP \\
\midrule
\textit{StaCon} & 161 & 33 (13.4\%) & 214 (86.6\%) \\
\textit{DynCon} & 144 & 56 (21.2\%) & 208 (78.8\%) \\
\textit{StaVar} & 53  & 33 (18.8\%) & 143 (81.2\%) \\
\textit{DynVar} & 56  & 34 (19.7\%) & 139 (80.3\%) \\ 
\bottomrule
\end{tabular}}
\end{table}

\reftab{tab:audio_results_synth} shows the auditory evaluation of the method when tested with synthetic sequences. The
remarkable achievement is the high number of right detections, around 80\%, in all cases. This means that neither the
dynamic character of the scene nor the fact that the number of AV objects varies have an impact on  sound detection. It
is also true that the number of false positives is large in all the cases.

\subsection{Results on Real Data}
\label{sec:results_real}
The \textit{Motion-Guided Robot Hearing} method was tested on the CTMS3 sequence of the CAVA data set \cite{Arnaud08}.
The CAVA (\textit{computational audio-visual analysis}) data set was specifically recorded to test various real-world
audio-visual scenarios. The CTMS3
sequence\footnote{\url{http://perception.inrialpes.fr/CAVA_Dataset/Site/data.html\#CTMS3}} consists on three
people freely moving in a room and taking speaking turns. Two of them count in English (one, two, three, ...) while the
third one counts in Chinese. The recorded signals, both auditory and visual, enclose the difficulties found in natural
situations. Hence, this is a very challenging sequence: People come in and out the visual field of the two cameras, hide
each other, etc. Aside from the speech sounds, there are acoustic reverberations and non-speech sounds such as those
emitted by foot steps and clothe chafing. Occasionally, two people speak simultaneously. 

\begin{figure}
\centering
  \begin{tabular}{ccc}
    \subfloat[]{\label{fig:ctms3_179}\includegraphics[width=0.3\linewidth]{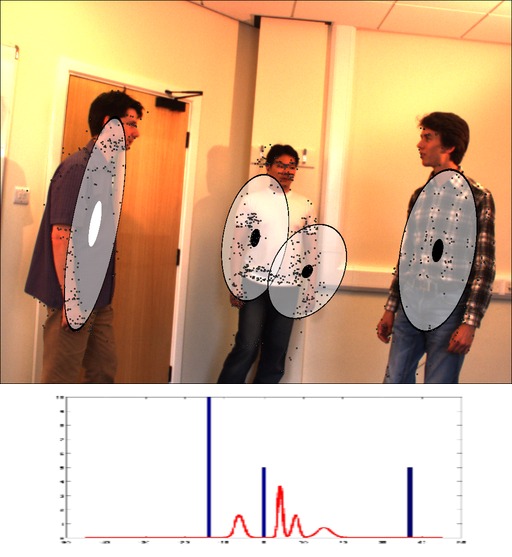}} \hspace{-0.3cm} &
    \subfloat[]{\label{fig:ctms3_222}\includegraphics[width=0.3\linewidth]{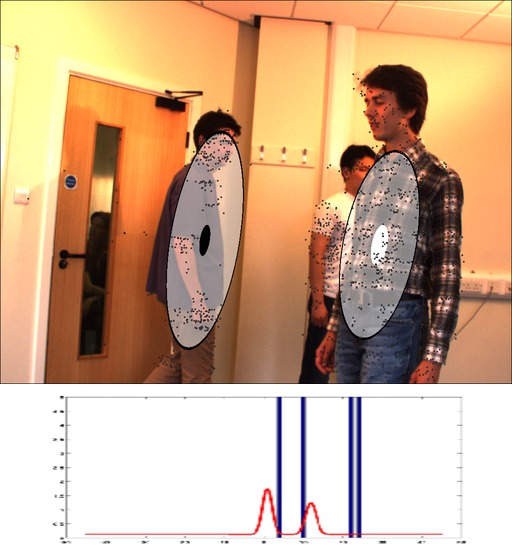}} \hspace{-0.3cm} &
    \subfloat[]{\label{fig:ctms3_254}\includegraphics[width=0.3\linewidth]{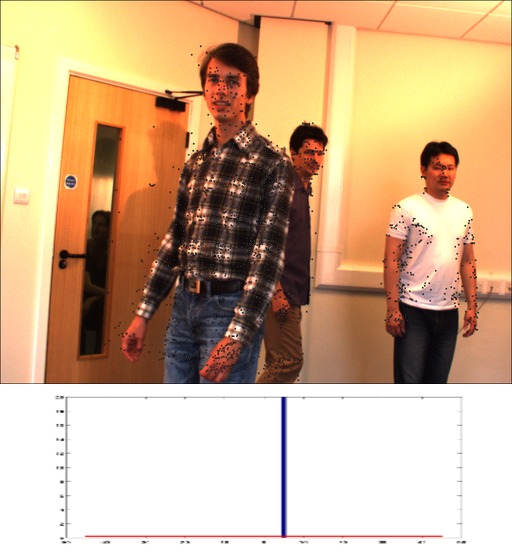}}  \\    
    \subfloat[]{\label{fig:ctms3_307}\includegraphics[width=0.3\linewidth]{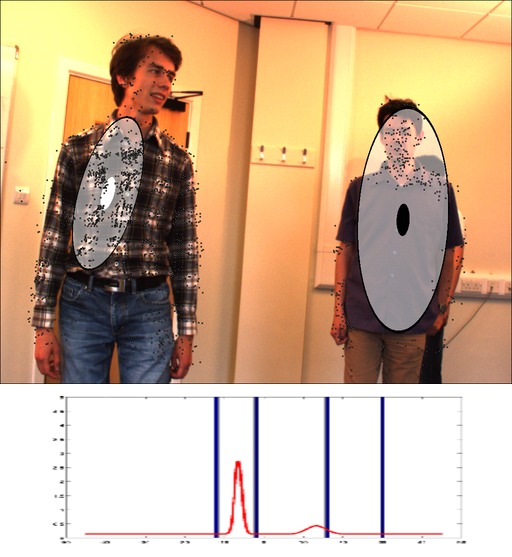}} \hspace{-0.3cm} &
    \subfloat[]{\label{fig:ctms3_259}\includegraphics[width=0.3\linewidth]{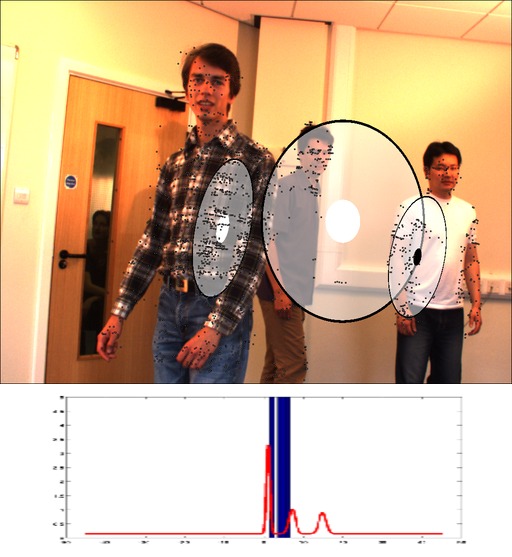}} \hspace{-0.3cm} &
    \subfloat[]{\label{fig:ctms3_297}\includegraphics[width=0.3\linewidth]{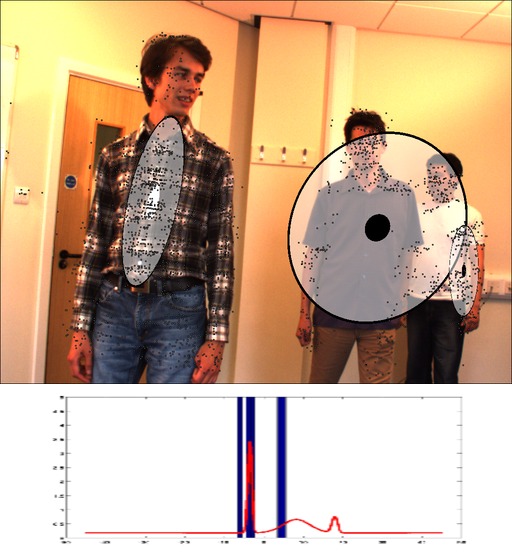}}  \\
    \subfloat[]{\label{fig:ctms3_349}\includegraphics[width=0.3\linewidth]{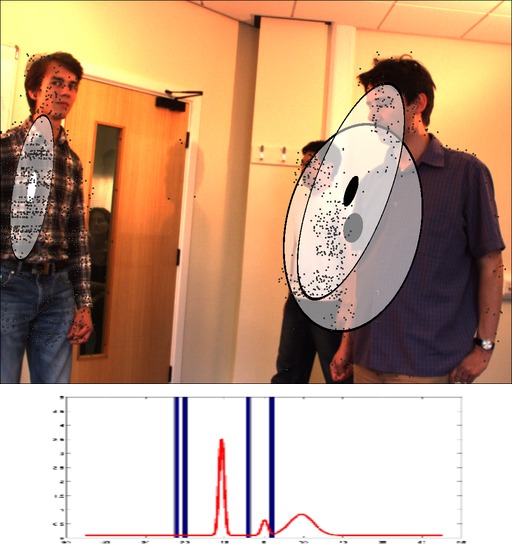}} \hspace{-0.3cm} &
    \subfloat[]{\label{fig:ctms3_191}\includegraphics[width=0.3\linewidth]{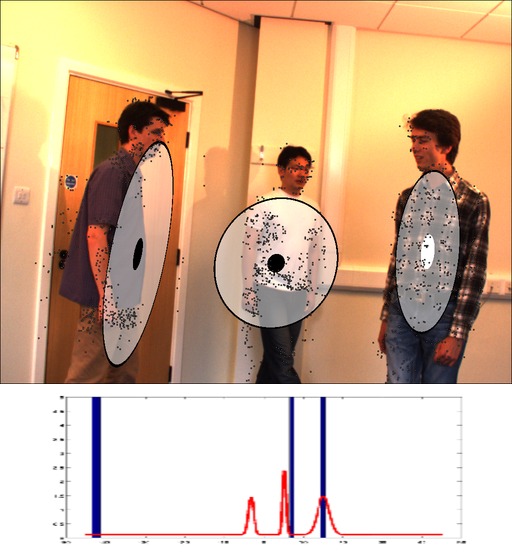}} \hspace{-0.3cm} &
    \subfloat[]{\label{fig:ctms3_204}\includegraphics[width=0.3\linewidth]{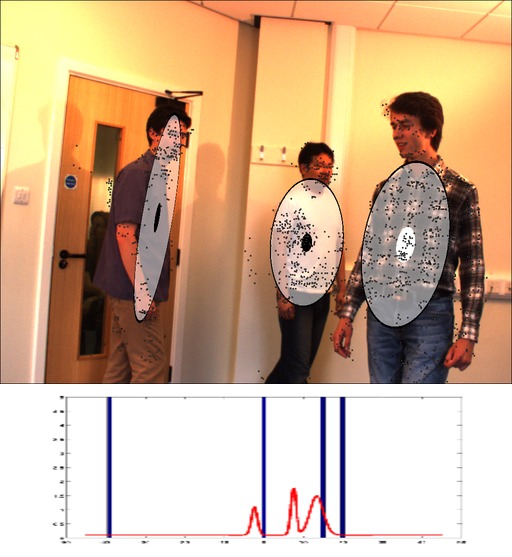}}  \\
  \end{tabular}
  \caption{Results obtained with the CTMS3 sequence from the CAVA data set. The ellipses correspond to the 3D covariance
matrices projected onto the image. The circle at each ellipse center illustrates the auditory activity: speaker emitting
a sound (white) or being silent (black) during each time interval. The plot associated with each image shows the
auditory observations as well as the fitted 1D mixture model.}
  \label{fig:results_ctms3}
\end{figure}

\reffig{fig:results_ctms3} shows the results obtained with nine time intervals chosen to show both successes and
failures of our method and to allow to qualitatively evaluate it. \reffig{fig:ctms3_179} shows one extreme case, in
which the distribution of the HM3D observations associated to the person with the white T-shirt is clearly not Gaussian.
\reffig{fig:ctms3_222} shows a failure of the \textit{ridgeline} method, used to merge Gaussian components, where
two different clusters are associated into one. \reffig{fig:ctms3_254} is an example with too few observations.
Indeed, the BIC points as optimal the model with no AV objects, thus considering all the observations to be outliers.
\reffig{fig:ctms3_307} clearly shows that our approach cannot deal with occluded objects, because of the
instantaneous processing of robocentric data, the person occluded will never be detected.
Figures~\ref{fig:ctms3_259},~\ref{fig:ctms3_297} and~\ref{fig:ctms3_349} are examples of success. The three speakers are
localised and their auditory status correctly guesses. However, the localisation accuracy is not good in these cases,
because one or more covariance matrices are not correctly estimated. The grouping of AV observations is, then, not well
conducted. Finally, Figures~\ref{fig:ctms3_191} and~\ref{fig:ctms3_204} show two case in which the \textit{Motion-Guided
Robot Hearing} algorithms works perfectly, three people are detected and their speaking activity is correctly assessed
from the ITD observations. In average, the method correctly detected 187 out of 213 objects (87.8\%) and correctly
detected the speaking state in 88 cases out of 147 (59.9\%).

\subsection{Results on NAO}
\label{sec:results_2}

To validate the \textit{Face-Guided Robot Hearing} method using NAO, we performed a set of experiments with five
different scenarios. The scenarios were recorded in a room around $5\times5$ meters with just a sofa and 3 chairs where
NAO and the other persons sat respectively. We designed five scenarios to test the algorithm in different conditions in
order to identify its limitations. Each scenario is repeated several times and consists on people counting from one up
to sixteen.

In scenario \textbf{S1}, only one person is in the room sitting in front of the robot and
counting. In the rest of the scenarios (\textbf{S2}-\textbf{S5}) three persons are in the room. People are not always in
the field of view (FoV) of the cameras and sometimes they move. In scenario \textbf{S2} three persons are sitting and
counting alternatively one after the other. The configuration of scenario \textbf{S3} is similar to the one of
\textbf{S2}, but one person is standing instead of sitting. These two scenarios are useful to determine the precision of
the ITDs and experimentally see if the difference of height (elevation) affects the quality of the extracted ITDs. The
scenario \textbf{S4} is different from \textbf{S2} and \textbf{S3} because one of the actors is outside the FoV. This
scenario is used to test if people speaking outside the FoV affect the performance of the algorithm. In the last
scenario (\textbf{S5}) the three people are in the FoV, but they count and speak independently of the other actors.
Furthermore, one of them is moving while speaking. With \textbf{S5}, we aim to test the robustness of the method to
dynamic scenes.

\begin{figure}
  \centering
  \begin{tabular}{cc}
\subfloat[\textbf{S1}]{\label{fig:s1}\includegraphics[width=0.45\linewidth]{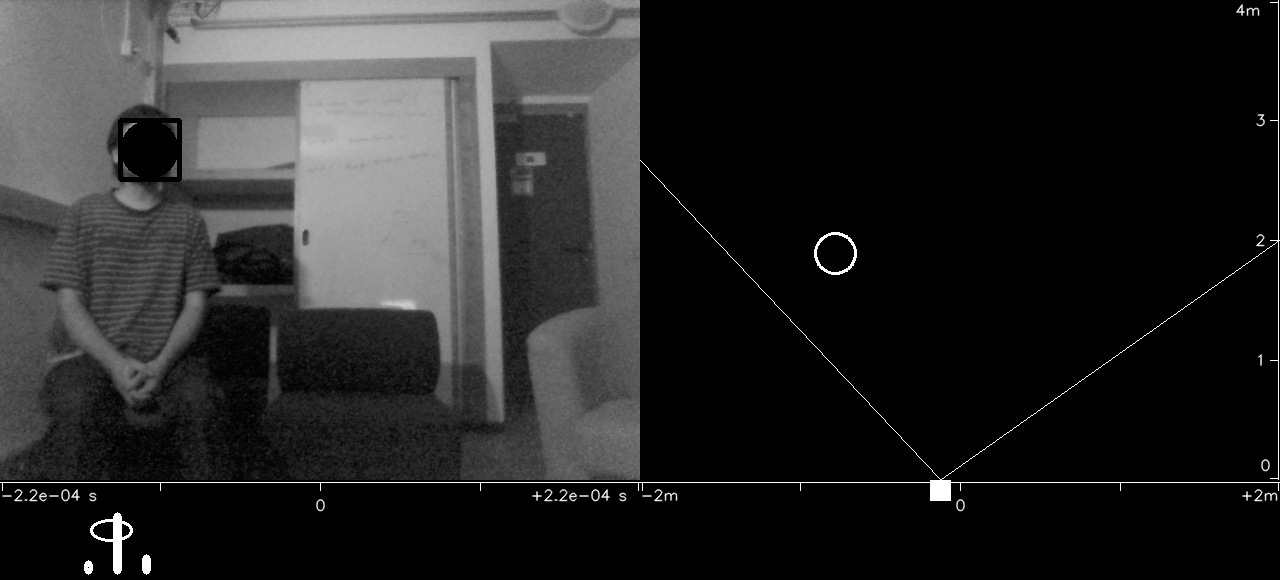}}
  & 
\subfloat[\textbf{S2}]{\label{fig:s21}\includegraphics[width=0.45\linewidth]{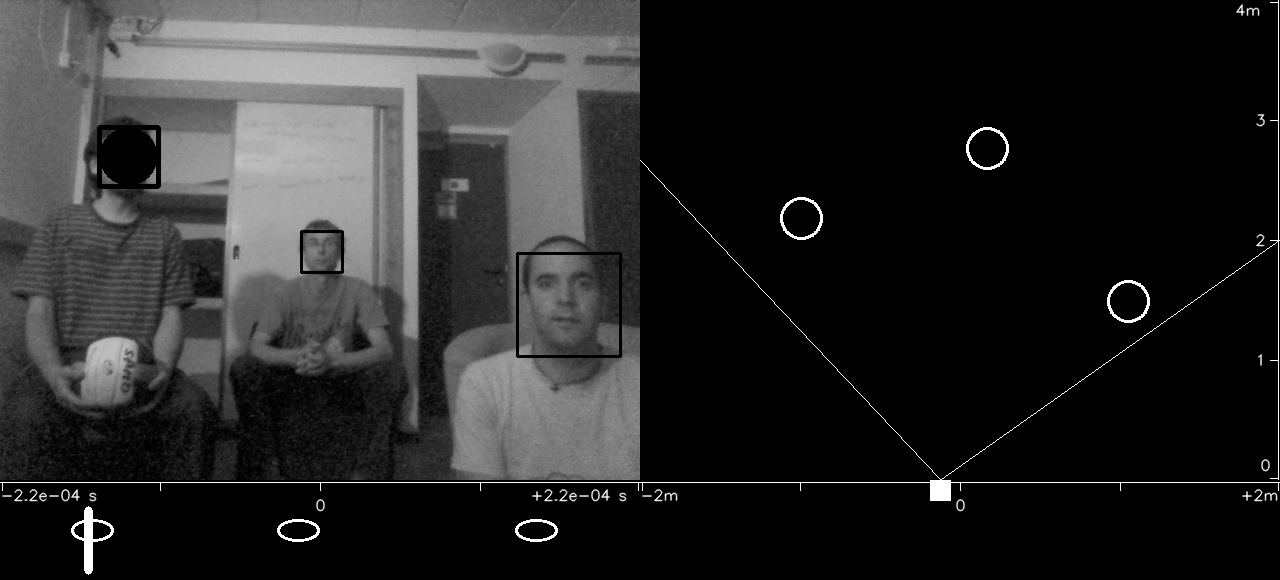}}
\\
\subfloat[\textbf{S4}]{\label{fig:s4}\includegraphics[width=0.45\linewidth]{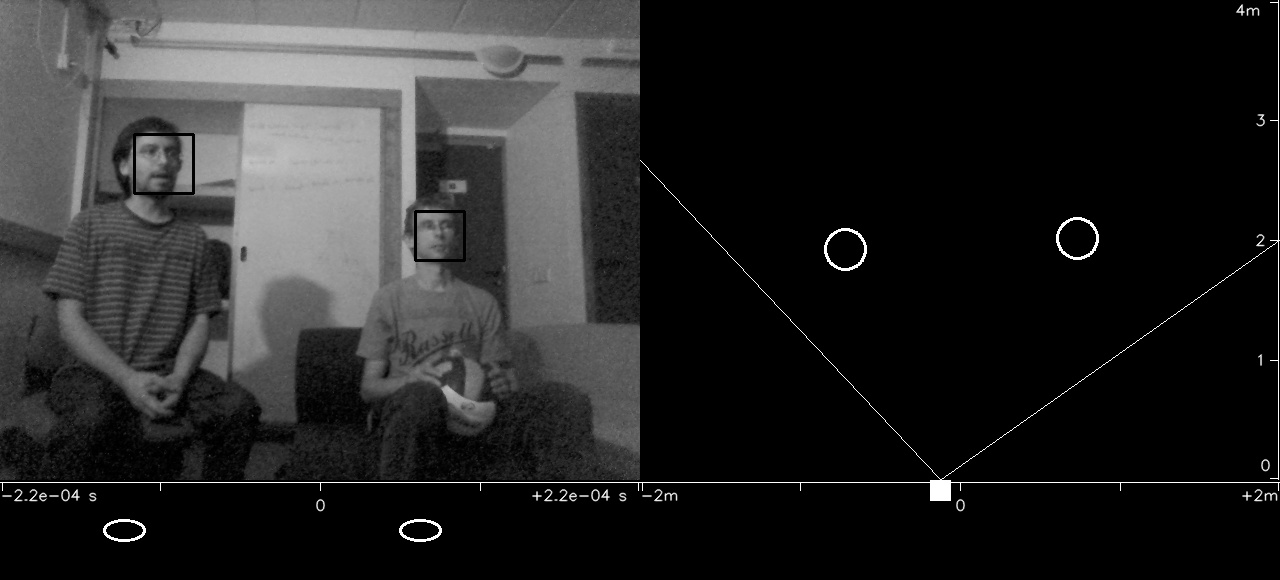}}
  &
\subfloat[\textbf{S5}]{\label{fig:s51}\includegraphics[width=0.45\linewidth]{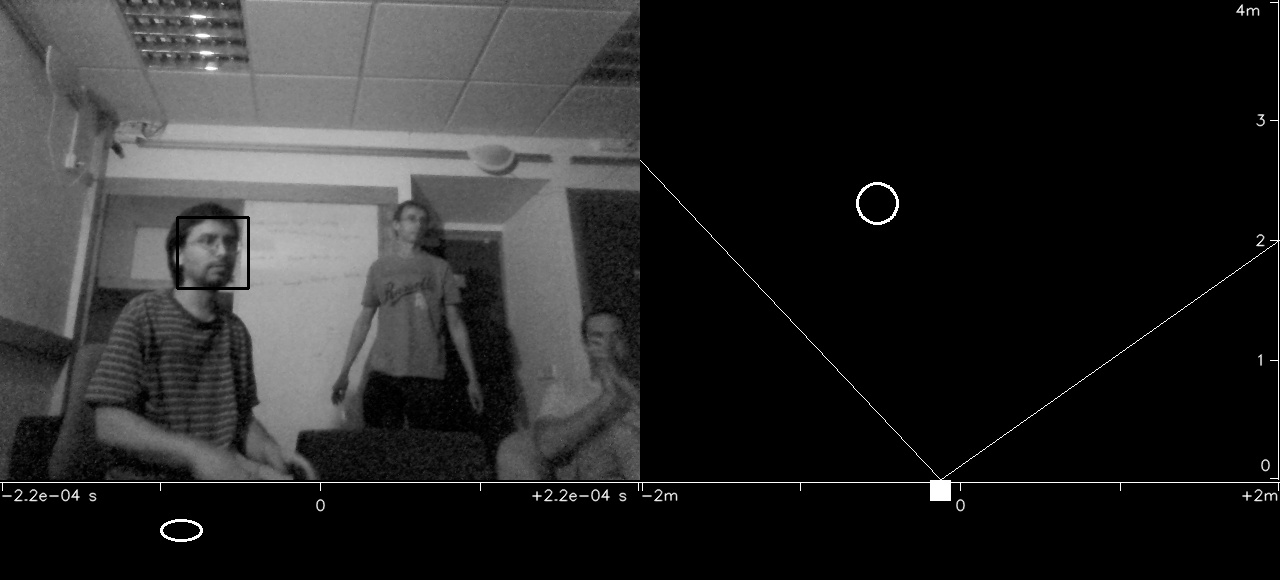}}
\\
\subfloat[\textbf{S5}]{\label{fig:s52}\includegraphics[width=0.45\linewidth]{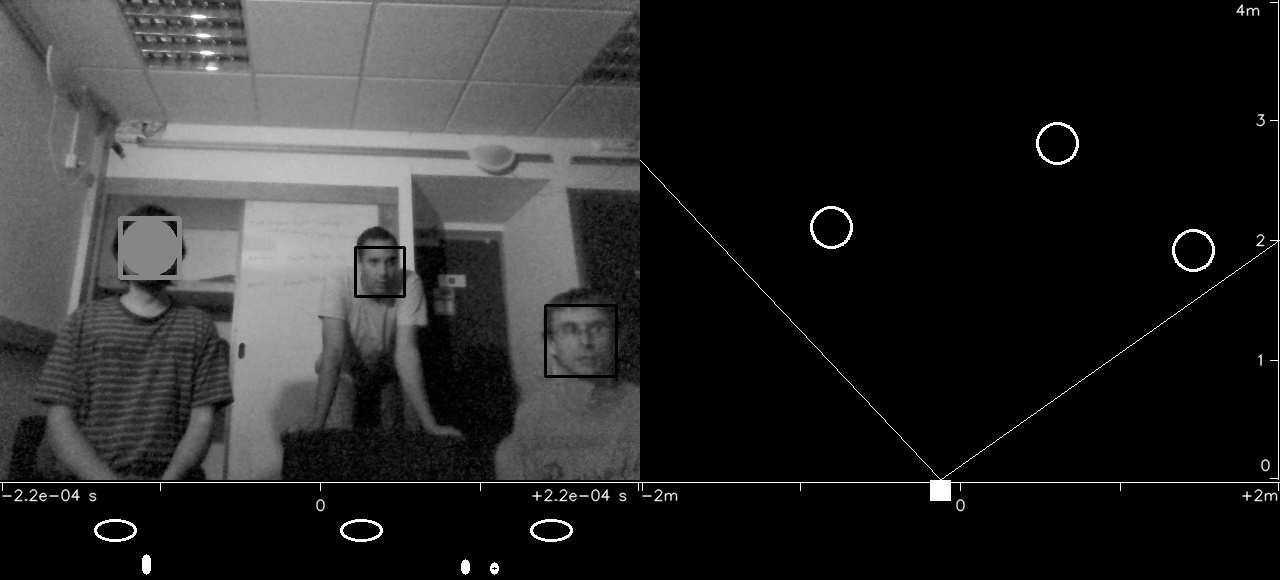}}
  &
\subfloat[\textbf{S2}]{\label{fig:s22}\includegraphics[width=0.45\linewidth]{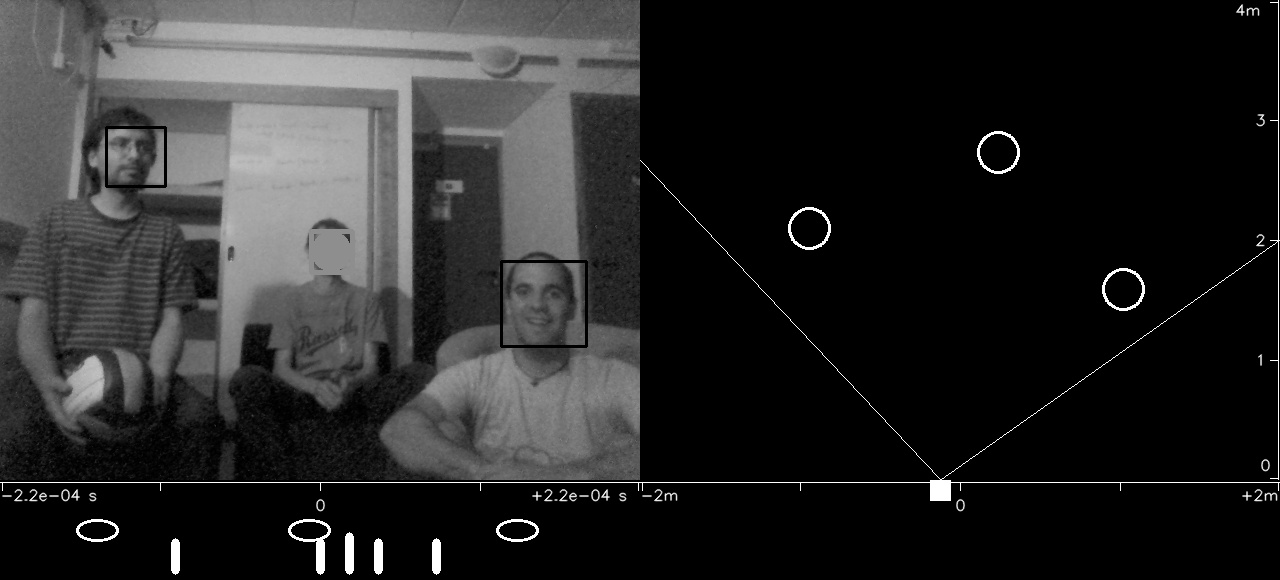}}
\\
\subfloat[\textbf{S3}]{\label{fig:s31}\includegraphics[width=0.45\linewidth]{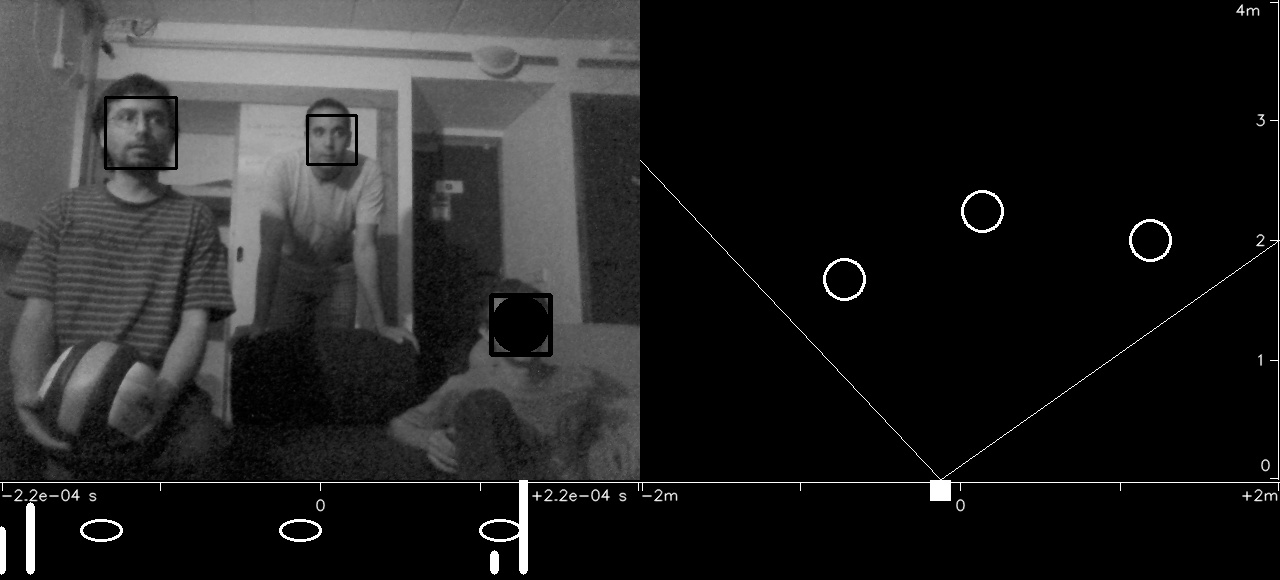}}
  &
\subfloat[\textbf{S3}]{\label{fig:s32}\includegraphics[width=0.45\linewidth]{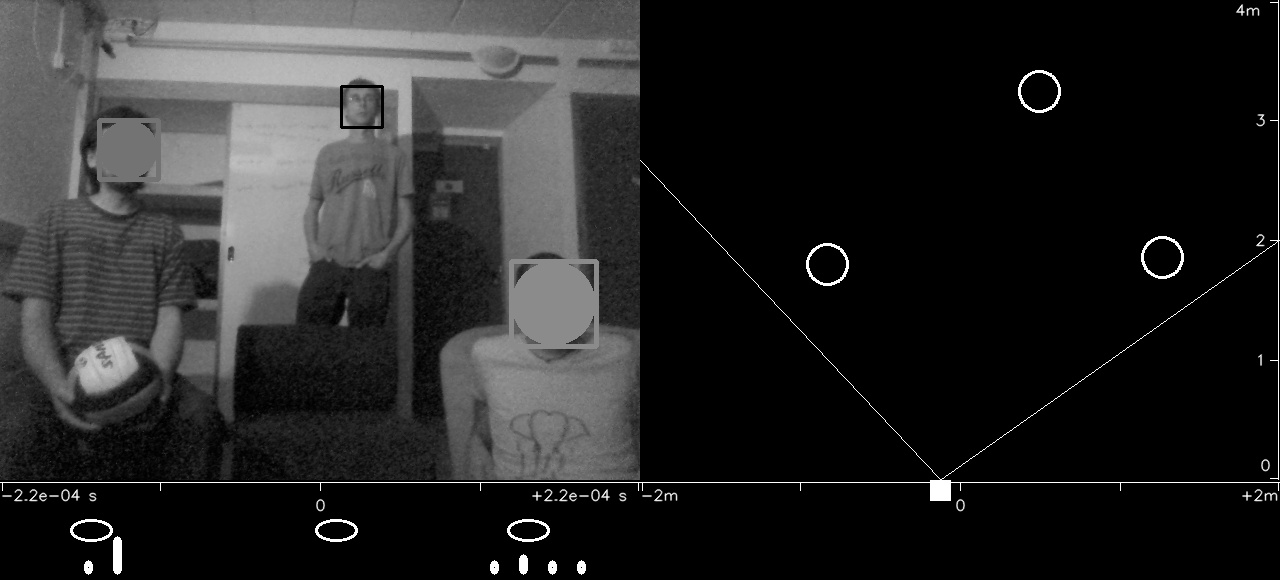}}
\\
  \end {tabular}
  \caption{Snapshots of the visualization tool. Frames selected among the five scenarios to show the method's strengths
and weaknesses. The faces' bounding box are shown superposed to the original image (top-left). The bird-view of the
scene is shown in the top-right part of each subimage. The histogram of ITD values as well as the projected faces are
shown in the bottom-left. See \refsec{sec:modular_structure} for how to interpret the images above.}
  \label{fig:results}
\end{figure}

In \reffig{fig:results} we show several snapshots of our visualization tool. These frames are selected from the
different scenarios aiming to show both the successes and the failures of the implemented system. \reffig{fig:s1} shows
an example of perfect alignment between the ITDs and the mapped face, leading to a high speaking probability. A similar
situation is presented in \reffig{fig:s21}, in which among the three people, only one speaks. A failure of the ITD
extractor is shown in \reffig{fig:s4}, where the actor in the left is speaking, but no ITDs are extracted. In
\reffig{fig:s51} we can see how the face detector does not work correctly: two faces are missing, one because of the
great distance between the robot and the speaker, and the other because it is partially out of the field of
view. \reffig{fig:s52} shows a snapshot of an AV-fusion failure, in which the extracted ITDs are not significant enough
to set a high speaking probability. The \reffig{fig:s22}, \reffig{fig:s31} and  \reffig{fig:s32} show the effect of
reverberations. While in \reffig{fig:s32} we see that the reverberations lead to the wrong conclusion that the actor
on the right is speaking, we also see that the statistical framework is able to handle
reverberations (\reffig{fig:s22} and \reffig{fig:s31}), hence demonstrating the robustness of the proposed
approach.

\reftab{tab:scenarios} shows the results obtained on scenarios (that were manually annotated). First of all we notice
the small amount of false negatives: the system misses very few speakers. A part from the first scenario (easy
conditions), we observe some false positives. These false positives are due to reverberations. Indeed, we notice how the
percentage of FP is severe in \textbf{S5}. This is due to the fact that high reverberant sounds (like hand claps) are
also present in the audio stream of this scenario. We believe that an ITD extraction method more robust to
reverberations will lead to more reliable ITD values, which in turn will lead to a better active speaker detector. It is
also worth to notice that actors in different elevations and non-visible actors do not affect the performance of the
proposed system, since the results obtained in scenarios \textbf{S2} to \textbf{S4} are comparable.

\begin{table}
\centering
{\small
\begin{tabular}{cccc}
\toprule
	      & FP & FN & TP\\
\midrule
  \textbf{S1} & 13 & 23 (13.4\%) & 149 (86.6\%)\\
  \textbf{S2} & 22 & 31 (14.9\%) & 176 (85.1\%)\\
  \textbf{S3} & 19 & 20 (11.3\%) & 157 (88.7\%)\\
  \textbf{S4} & 37 & 12 (6.7\%)  & 166 (93.3\%)\\
  \textbf{S5} & 53 & 32 (19.0\%) & 136 (81.0\%)\\ 
\bottomrule
\end{tabular}}
\caption{Quantitative evaluation of the proposed approach for the five scenarios. The columns represent, in order: the
amount of correct detections (CD), the amount of false positives (FP), the amount of false negatives (FN) and the
total number of counts (Total).}
\label{tab:scenarios}
\end{table}

\section{Conclusions and Future Work}
\label{sec:conc}

This paper introduces a multimodal hybrid probabilistic/deterministic framework for simultaneous detection and
localization of speakers. On one hand, the deterministic component takes advantage of the geometric and physical
properties associated with the visual and auditory sensors: the audio-visual mapping $(\overline{\cal A}\circ{\cal V})$
allows us to transform the visual features from the 3D space to an 1D auditory space. On the other hand,
the probabilistic model deals with the observation-to-speaker assignments, the noise and the outliers. We propose a new
multimodal clustering algorithm based on a 1D Gaussian mixture model, an initialization procedure, and a model selection
procedure based on the BIC score. The method is validated on a humanoid robot and interfaced through the RSB middleware
leading to a platform-independent implementation.

The main novelty of the approach is the visual guidance. Indeed, we derived to EM-based procedures for
\textit{Motion-Guided} and \textit{Face-Guided} robot hearing. Both algorithms provide the number of speakers,
localize them and ascertain their speaking status. In other words, we show how one of the two modalities can be
used to supervise the clustering process. This is possible thanks to the audio-visual calibration procedure that
provides an accurate projection mapping $(\overline{\cal A}\circ{\cal V})$. The calibration is specifically designed
for robotic usage since it requires very few data, it is long-lasting and environment-independent.

The presented method solves several open methodological issues: (i) it fuses and clusters visual and auditory
observations that lie in physically different spaces with different dimensionality, (ii) it models and estimates the
object-to-observation assignments that are not known, (iii) it handles noise and outliers mixed with both visual and
auditory observations whose statistical properties change across modalities, (iv) it weights the relative importance of
the two types of data, (v) it estimates the number of AV objects that are effectively present in the scene during a
short time interval and (vi) it gauges the position and speaking state of the potential speakers.

One prominent feature of our algorithm is its robustness. It can deal with various kinds of perturbations, such as the
noise and outlier encountered in unrestricted physical spaces. We illustrated the effectiveness and robustness of our
algorithm using challenging audio-visual sequences from a publicly available data set as well as using the humanoid
robot NAO in regular indoor environments. We demonstrated good performance on different scenarios involving
several actors, moving actors and non-visible actors. Interfaced by means of the RSB middleware, the
\textit{Face-Guided Robot Hearing} method processes the audio-visual data flow from two microphones mounted inside the
head of a companion robot with noisy fans and two cameras at a rate of 17~Hz.

There are several possible ways to improve and to extend our method. Our current implementation relies more on the
visual data than on the auditory data, although there are many situations where the auditory data are more reliable. The
problem of how to weight the relative importance of the two modalities is under investigation. Our algorithm can also
accommodate other types of visual cues, such as 2D or 3D optical flow, body detectors, etc., or auditory cues, such as
Interaural Level Differences. In this paper we used one pair of microphones, but the method can be easily extended to
several microphone pairs. Each microphone pair yields one ITD space and combining these 1D spaces would provide a much
more robust algorithm. Finally, another interesting direction of research is to design a dynamic model
that would allow to initialize the parameters in one time interval based on the information extracted in several
previous time intervals. Such a model would necessarily involve dynamic model selection, and would certainly help to
guess the right number of AV objects, particularly in situations where a cluster is occluded but still in the visual
scene, or a speaker is highly interfered by another speaker/sound source. Moreover, this future dynamic model selection
should be extended to provide for audio-visual tracking capabilities, since they enhance the temporal coherence of the
perceived audio-visual scene.

\section*{Acknowledgments}
This work was partially funded by the HUMAVIPS FP7 European Project FP7-ICT-247525.

\bibliographystyle{these}
\bibliography{refs.bib}
\end{document}

%% file: introduction-radu.tex

For the last decade, robotics research has developed the concept of human companions endowed with cognitive skills and
acting in complex and unconstrained environments. While a robot must still be able to safely navigate and manipulate
objects, it should also be able to interact with people. Obviously, speech communication plays a crucial role in
modeling the cognitive behaviors of robots. But in typical real-world scenarios, humans that emit speech (as well as other sounds of
interest) are at some distance and hence the robot's microphone signals are strongly impaired by noise, reverberations,
and interfering sound sources. Compared with other types of hands-free human-machine audio interfaces, e.g., smart
phones, the human to robot distance is larger. Moreover, the problem is aggravated further as the robot produces
significant \textit{ego noise} due to its mechanical drives and electronics. This implies that robot-embodied cognition
cannot fully exploit state-of-the-art speech recognition and more generally human-robot interaction based on verbal
communication. 

Humans have sophisticated abilities to enhance and disambiguate weak unimodal data based on information fusion from
multiple sensory inputs \cite{Anastasio00,King09}. In particular, audio-visual fusion is one of the most
prominent forms of multimodal data processing and interpretation mechanisms; it plays a crucial role in extracting
auditory information from dirty acoustic signals \cite{Haykin05}. In this paper we address the problem of how
to detect and localize people that are both seen and heard by a humanoid robot. We are particularly interested in
combining vision and hearing in order to identify the activity of people, e.g., emitting speech and non-speech sounds,
in informal scenarios and complex visuo-acoustic environments. 

\begin{figure}[t!]
 \centering
\includegraphics[width=0.7\linewidth]{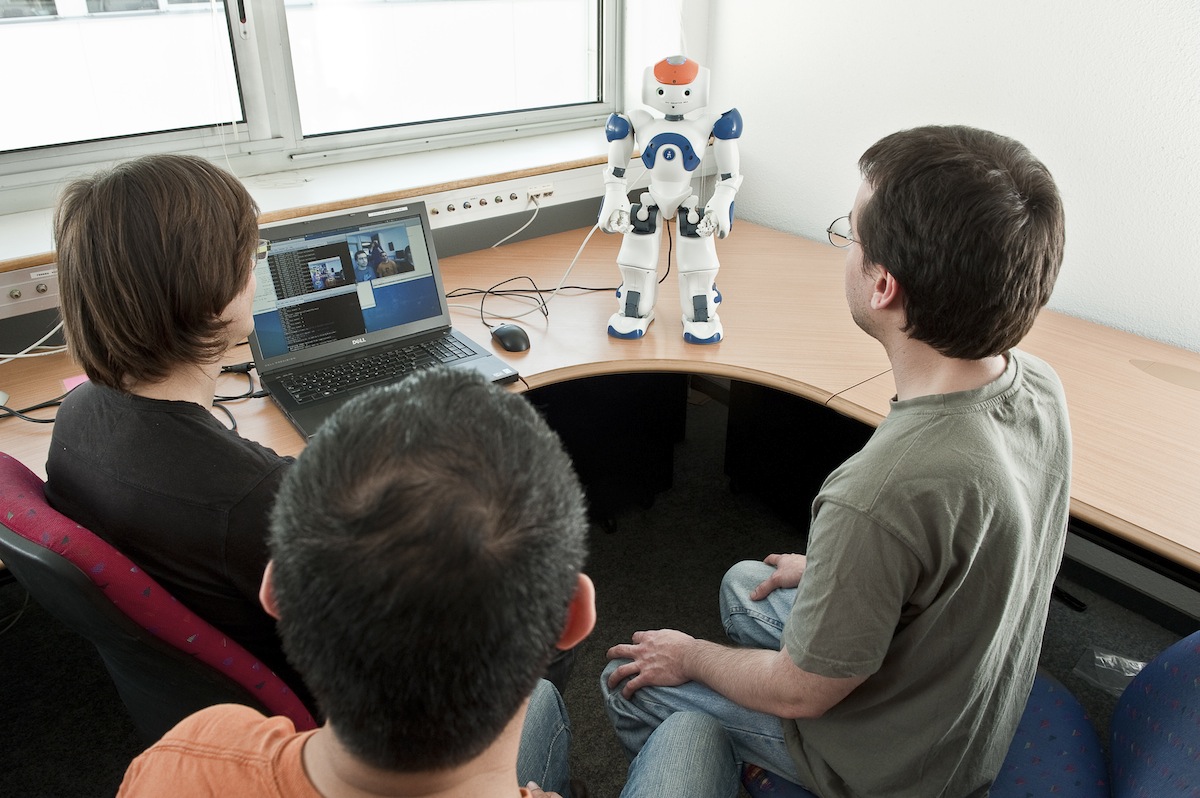}
\caption{A typical scenario in which a companion humanoid robot (NAO) performs audio-visual fusion in an attempt to
assess the auditory status of each one of the speakers in front of the robot and to estimate the 3D locations of their
faces. The method uses the robot's onboard cameras and microphones as well as a modular software architecture based on
the freely available RSB (robotics service bus) middleware. This allows \textit{untethered interaction} between robots
and people. Moreover, RSB allows remote algorithm implementation using external computing power and without the
performance and deployment restrictions imposed by the onboard computing resources.
}
 \label{fig:scenario_intro}
\end{figure}

A typical example of such a scenario is shown in \reffig{fig:scenario_intro} where people sit at some distance from
the robot and informally chat which each other and with the robot. The robot's first task (prior to speech recognition,
language understanding, and dialog handling) consists in retrieving the time-varying auditory status of the speakers.
This allows the robot to turn its attention towards an acoustically active person, precisely determine the position and
orientation of its face, optimize the emitter-to-receiver acoustic pathway such as to maximize the signal-to-noise
ratio (SNR), and eventually retrieve a clean speech signal. We note that this problem cannot be solved within the
traditional human-computer interface paradigm which is based on \textit{tethered} interaction, i.e., the user wears a
close-range microphone, and which primarily works for a single user and with clean acoustic data. On the contrary,
untethered interaction is likely to broaden the range of potential cooperative tasks between robots and people, to allow
natural behaviors, and to enable multi-party dialog. 

This paper has the following two main contributions:
\begin{itemize}

\item The problem of detection and localization of multiple audio-visual (AV) events is cast into a mixture model. We
explore the emitter-to-perceiver acoustic propagation model that allows us to map \textit{both} 3D visual features and
3D sound sources onto the 1D auditory space spanned by interaural time differences (ITD) between two microphones.
Therefore, visual and auditory cues can be clustered together to form AV events. We derive an expectation-maximization
(EM) procedure that exhibits several interesting features: it allows either to put vision and hearing on an equal
footing, or to weight their relative importance such that the algorithm can be partially supervised by the most reliable
of the two modalities, it allows to perform model selection or, more precisely, to estimate the number of AV events, it
is robust to outliers, such as visual artifacts and reverberations, it is extremely efficient as it relies on a
one-dimensional Gaussian mixture model and as the 3D event locations can be inferred without any additional effort. 

\item The proposed model and method are implemented in real-time using a stereoscopic camera pair and two microphones
embedded into the head of the humanoid companion robot NAO, manufactured by Aldebaran Robotics. We describe a modular
software architecture based on the freely available \textit{Robotics Service Bus} (RSB) middleware. RSB events are equipped with
several timestamps, thus handling the synchronization of visual and auditory observations gathered at different sampling
rates as well as the synchronization of higher level visual and auditory processing modules. This software architecture
allows to implement and test our algorithms remotely without the performance and deployment restrictions imposed by the
robot platform itself. More interestingly, the proposed implementation can be reused with other robots.

\end{itemize}

The remainder of the paper is organized as follows: \refsec{sec:sota} delineates the related published work,
\refsec{sec:problem} outlines the hybrid deterministic/probabilistic model, \refsec{sec:features} gives the details of
the auditory and visual extracted features, Sections~\ref{sec:inference} and~\ref{sec:robot_implem} describe the
multimodal inference procedure as well as its on-line implementation on the humanoid robot NAO, \refsec{sec:results}
shows the results we obtained and \refsec{sec:conc} draws some conclusions and future work guidelines.